\documentclass[12pt,a4paper]{ozu-thesis}

\usepackage{amsmath,amssymb,amsthm}
\usepackage{bm} 
\usepackage{mathtools} 
\usepackage{graphicx} 
\usepackage{svg} 
\usepackage{booktabs} 
\usepackage{rotating} 
\usepackage{longtable} 
\usepackage{multirow} 
\usepackage{makecell} 
\usepackage{colortbl} 
\usepackage{threeparttable} 
\usepackage{adjustbox} 
\usepackage{float} 
\usepackage{placeins} 
\usepackage{afterpage} 
\usepackage{tikz} 
\usepackage{caption} 
\usepackage{subcaption} 
\usepackage{indentfirst} 
\usepackage[printonlyused]{acronym} 
\usepackage{csquotes} 
\usepackage{algorithm} 
\usepackage{algorithmic} 
\usepackage{listings} 
\usepackage{xcolor} 
\usepackage{url} 
\usepackage{xurl} 
\usepackage{comment} 
\usepackage{type1cm} 
\usepackage{anyfontsize} 
\usepackage{siunitx} 
\usepackage{cleveref} 
\usepackage{pifont}
\usepackage[titles]{tocloft}

\DeclareCaptionFormat{custom}
{%
   \textbf{#1#2}\textit{\normalsize #3}
}
\definecolor{codegreen}{rgb}{0,0.6,0}
\definecolor{codegray}{rgb}{0.5,0.5,0.5}
\definecolor{codepurple}{rgb}{0.58,0,0.82}
\definecolor{backcolour}{rgb}{0.95,0.95,0.92}
\lstdefinestyle{codestyle}{
    backgroundcolor=\color{backcolour},   
    commentstyle=\color{codegreen},
    keywordstyle=\color{magenta},
    numberstyle=\tiny\color{codegray},
    stringstyle=\color{codepurple},
    basicstyle=\ttfamily\footnotesize,
    breakatwhitespace=false,         
    breaklines=true,                 
    captionpos=b,                    
    keepspaces=true,                 
    numbers=left,                    
    numbersep=5pt,                  
    showspaces=false,                
    showstringspaces=false,
    showtabs=false,                  
    tabsize=2
}

\theoremstyle{definition}

\title{GaussianOcc3D: Gaussian Based Adaptive Multi-Modal 3D Semantic Occupancy Prediction}

\author{Abdullah Enes Doruk}
\department{Department Of Artificial Intelligence and Data Engineering}
\faculty{Graduate School of Science and Engineering }

\copyrightyear{2025}
\submitdate{December, 2025}
                         
\approveddate{December 30, 2025}

\degree{Master of Science}

\mainadvisor{Prof. Hasan Fehmi Ates}

\def\possibility{3}

\if\possibility1

\coadvisor{}
\firstmember{Full Title Name Surname, Advisor}[Department of ...]
\secondmember{Full Title Name Surname} [Department of ...] 
\thirdmember{Full Title Name Surname} [Department of ...][... University]

\else
\if\possibility2

\firstmember{Full Title Name Surname, Advisor}[Department of ...]
\secondmember{Full Title Name Surname, Co-advisor} [Department of ...] [... University]
\thirdmember{Full Title Name Surname} [Department of ...] [...  University]
\fourthmember{Full Title Name Surname}[Department of ...][... University]
\fifthmember{Full Title Name Surname}[Department of ...] [... University]

\else

\coadvisor{Full Title Name Surname}
\coadvisortrue

\firstmember{Full Title Name Surname, Advisor}[Department of ...]
\secondmember{Full Title Name Surname} [Department of ...] 
\thirdmember{Full Title Name Surname} [Department of ...] [... University]

\fi
\fi

\bibfiles{bibliography}

\signaturepagefalse
\copyrightfalse
\figurespagefalse
\tablespagefalse

\begin{document}
\bibliographystyle{ieeetr}
\begin{preliminary}

\begin{dedication}
\null\vfil
{\large
\begin{center}
\noindent To everyone who believed in me when I doubted myself, I am deeply grateful. And above all, thanks to Almighty God.
\end{center}}
\vfil\null
\end{dedication}


\begin{abstract}
\noindent
The sparse object detection paradigm shift towards dense 3D semantic occupancy prediction is necessary for dealing with long-tail safety challenges for autonomous vehicles. Nonetheless, the current voxelization methods commonly suffer from excessive computation complexity demands, where the fusion process is brittle, static, and breaks down under dynamic environmental settings. To this end, this research work enhances a novel Gaussian-based adaptive camera-LiDAR multimodal 3D occupancy prediction model that seamlessly bridges the semantic strengths of camera modality with the geometric strengths of LiDAR modality through a memory-efficient 3D Gaussian model. The proposed solution has four key components: (1) LiDAR Depth Feature Aggregation (LDFA), where depth-wise deformable sampling is employed for dealing with geometric sparsity, (2) Entropy-Based Feature Smoothing, where cross-entropy is employed for handling domain-specific noise, (3) Adaptive Camera-LiDAR Fusion, where dynamic recalibration of sensor outputs is performed based on model outputs, and (4) Gauss-Mamba Head that uses Selective State Space Models for global context decoding that enjoys linear computation complexity. Experiments performed for thorough validation of the proposed solution against the OpenOccupancy, Occ3D-nuScenes, and SemanticKITTI benchmarks show that this solution achieves of 25.3\%, 49.4\%, and 25.2\%, respectively, for mIoU, clearly establishing a significant improvement over current solutions for finer geometric reconstruction such as thin objects, vegetation, under diverse lighting, and weather conditions.

\textbf{\textit{Keywords}}: 3D Occupancy prediction, Multi-modal learning, Gaussian representation

\end{abstract}
\begin{ozetce}
\noindent
Semantik nesne tespitine dayalı yaklaşımlardan yoğun 3B anlamsal doluluk (semantic occupancy) tahminine geçiş, otonom araçlarda uzun kuyruklu güvenlik sorunlarının ele alınabilmesi açısından zorunlu hâle gelmiştir; ancak mevcut voksel tabanlı yöntemler yüksek hesaplama maliyetleriyle sınırlanmakta ve kullanılan füzyon mekanizmaları çoğunlukla kırılgan, statik ve dinamik çevresel koşullar altında yetersiz kalmaktadır. Bu çalışmada, kamera modalitesinin anlamsal zenginliğini LiDAR modalitesinin geometrik doğruluğu ile bellek açısından verimli bir 3B Gauss temsili üzerinden bütünleştiren, Gaussian tabanlı uyarlanabilir kamera–LiDAR çok-modlu 3B doluluk tahmin modeli geliştirilmektedir. Önerilen yöntem; geometrik seyrekliği gidermek amacıyla derinlik eksenli deformable örnekleme kullanan LiDAR Derinlik Özellik Birleştirme (LDFA) modülünü, alan-özgül gürültüyü bastırmak için çapraz entropi temelli Entropi Tabanlı Özellik Yumuşatma mekanizmasını, model çıktılarındaki belirsizliğe bağlı olarak sensör katkılarını dinamik biçimde yeniden ağırlıklandıran Uyarlanabilir Kamera–LiDAR Füzyonu stratejisini ve doğrusal hesaplama karmaşıklığına sahip Seçici Durum Uzay Modelleri ile küresel bağlamı etkin şekilde çözen Gauss-Mamba Başlığını içermektedir. OpenOccupancy, Occ3D-nuScenes ve SemanticKITTI veri kümeleri üzerinde gerçekleştirilen kapsamlı deneyler, önerilen yaklaşımın sırasıyla \%25.3, \%49.4 ve \%25.2 mIoU değerleriyle mevcut yöntemlere kıyasla yeni son-teknik sonuçlar elde ettiğini göstermekte; özellikle ince yapılı nesnelerin, bitki örtüsünün ve farklı aydınlatma ile hava koşulları altındaki sahnelerin daha hassas geometrik yeniden inşasında belirgin üstünlük sağladığını ortaya koymaktadır.

\textbf{\textit{Anahtar Kelimeler:}} 3B Doluluk tahmini, Çok modlu öğrenme, Gauss temsili

\end{ozetce}
\begin{acknowledgements}
I would like to express my deepest gratitude to my supervisor, Prof. Dr. Hasan Fehmi Ates, for his invaluable guidance, continuous support, and insightful feedback throughout the course of this thesis. Special thanks to my colleagues and friends at DEEP-VIP Laboratory/Ozyegin University  for their support and encouragement. Finally, I would like to thank my family for their unconditional love and belief in me.
\end{acknowledgements}

\contents

%
\end{preliminary}

\chapter{Introduction}

The development of autonomous driving (AD) is an immensely transformative technological effort underway in today’s world that has remarkably progressed from basic assist and semi-autonomous systems to fully self-driving AI models that can navigate complex human environments autonomously. This technological shift promises to revolutionize the manner of transport within various segments such as urban robotaxi solutions, freight and long-haul transport solutions, last-mile delivery solutions, and personal transport solutions mainly through error elimination and increased efficiency. Fundamentally speaking, an autonomous vehicle is able to navigate through environments through an advanced and complex sense-plan-act cycle that includes collecting raw sensor data from various sources of sensor input such as LiDAR, camera sensors, and radar sensors. Among various sensor inputs and operations involved within an automatic vehicle, the perception layer is the safety backbone that bears the prime task of extracting relevant semantic meaning from raw sensor data~\cite{GenericADSurvey, xu2025survey}.

Historically, this perception task has been posed as the problem of 3D object detection, where the task is to approximate dynamic agents (such as vehicles and pedestrians) using discrete 3D bounding boxes. Although this object-centric view has propelled the early adoption of AD systems, this scheme has traditionally posed a very simplistic view towards physical reality by drastically reducing the rich geometry of the environment to cuboid shapes. By design, this view has inherently failed to capture general objects that defy classification into predefined categories, such as irregular pieces of construction trash, vegetation overhangs, curved barriers, as well as roadway hazards with unpredictable shapes. Therefore, this sparse object detection alone has introduced a perception blind spot into AD systems that call for an urgent need to move towards a dense, fine-grained understanding of the 3D scene that characterizes both occupied space and free space with high geometric accuracy~\cite{GenericADSurvey, xu2025survey}.

\begin{figure}[ht]
    \setlength{\belowcaptionskip}{0pt}
    \centering
        \includegraphics[width=1\textwidth]{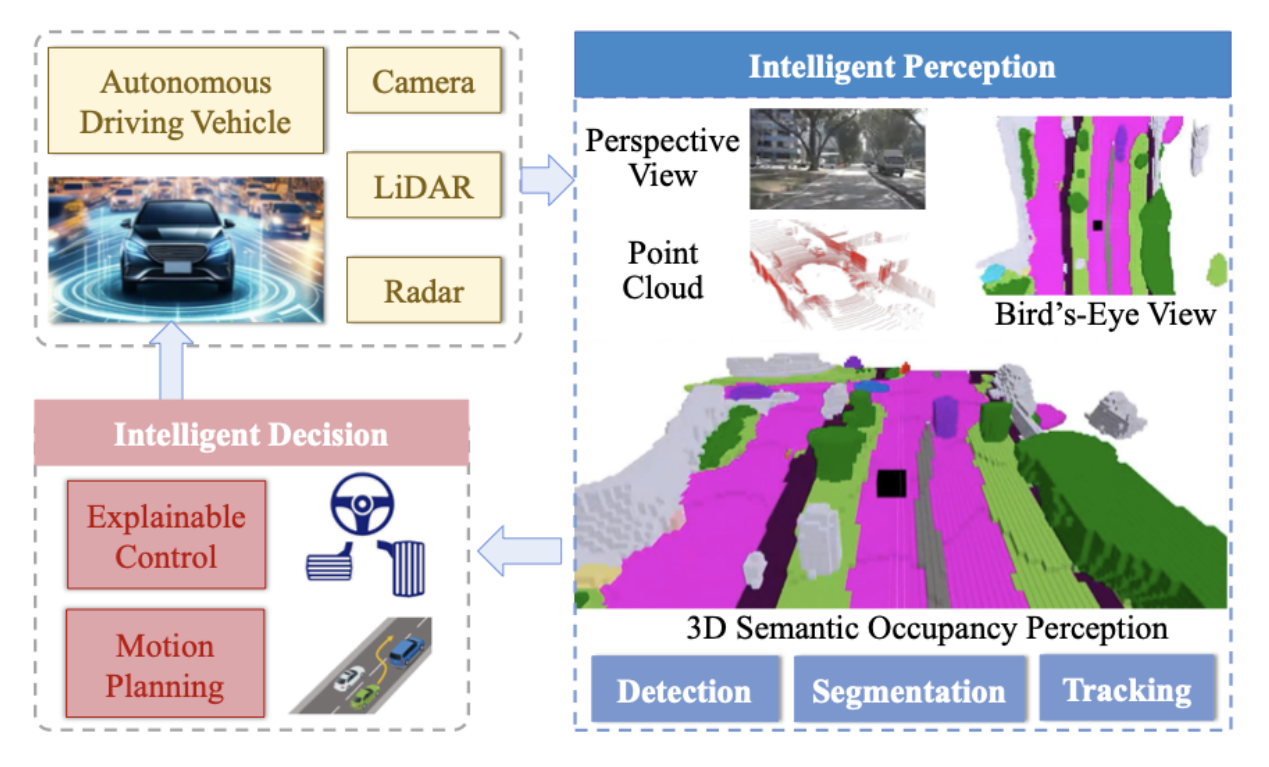}
    \label{fig:general}
\end{figure}

\section{The Paradigm Shift to 3D Occupancy}

In response to the inherent challenges of sparse object detection, the state of the art of autonomous perception has witnessed the advent of the 3D Semantic Occupancy Prediction paradigm. This paradigm change has been further fueled by its application in industrial-scale projects, with special focus on the application during theTesla AI Day talks of the year 2022 \cite{tesla2022aiday}. The change also satisfies the inherent requirement of dealing with the long-tail problem in the context of autonomous perception—the detection of unusual, irregular, and/or undefined road hazards (e.g., overturned vehicles, fallen rubble, and/or construction overhangs) that defy the traditional rigid ontology of the conventional detectors’ predefined notions of object type \cite{xu2025survey, GenericADSurvey}. In contrast to the traditional notion of object detection in the 3D domain, wherein the aim is to demarcate physical objects as convex bounding boxes—losing geometric detail and failing to accommodate geometric irregularities for undefined road hazards—the semantic occupancy prediction approach discretizes the entire 3D space of the scenario into a volumetric density grid that is contiguous and compact. In this context, the perception problem formulation deviates significantly, as the system now requires the assignment of a geometric state (occupied/free) as well as a semantic label to each elementary unit of the volumetric density grid itself \cite{xu2025survey, GenericADSurvey}. This fundamental change in the nature of perception problems naturally brings about the dualistic advantage of bounding boxes limited to the detection of predefined semantic object classes, as opposed to the more general occupational grid detectors having the capability to define the geometric shape of any physically occurring structure, irrespective of predefined semantic mappings. By demarcating the boundary between the free space and the occupied matter of the environment, this volumetric density representation provides the needed granularity of environmental awareness for effective and safe path methodology formulation for self-driving vehicles operating on unstructured environments when cuboid geometry approximations would be grossly inadequate \cite{xu2025survey, GenericADSurvey}.

\section{Sensory Modalities and Limitations}


The veracity of any 3D perception system will be always limited by the inherent nature of the sensing components and the underlying data representation. In the realm of Computer Vision, the camera serves as a passive photometric sensor that records a high-frequency sampling of the signal array rich in color and texture \cite{xu2025survey, GenericADSurvey}. This semantic richness is critical for aesthetic and semantic image understanding tasks that allow the system to identify fine-grained textual clues, infer the state of a traffic signal, and identify semantically similar classes like road markings and dividing lanes. Unfortunately, as the domain moves towards the paradigm of 3D Vision, the monocular vision system faces a strict mathematical limit—a two-dimensional image contains a non-injective mapping from the three-dimensional scene to a two-dimensional image plane that clearly classifies Depth Estimation to be an ill-posed problem. Without resorts to direct Time-of-Flight \cite{realtimeocc2024, liu2023lidar4d} techniques on the image plane components, FB-OCC \cite{li2023fbocc} and OccDepth \cite{miao2023occdepth} solutions must be based on learned depth priors that are prone to be severely misled in repetitive texture regions and poor ambient lighting conditions that cause glare from high dynamic range.

\sloppy
\section{Multi-Modal Synergy: Representation and Fusion Strategies}

The key to advances in strong 3D occupancy prediction lies in the principle that a necessary merge of semantic understanding and geometric accuracy cannot be accomplished by any one modality alone. The motivations for integrating camera and LiDAR modalities are based on the observation that their characteristics are strictly complementary. On one hand, the camera provides a dense photometric channel (color and texture) that is required to disambiguate semantic interpretations, such as distinguishing between a road surface suitable for driving and a flat concrete sidewalk, whereas on the other hand, the LiDAR provides a sparse high-accuracy geometric manifold that informs scene geometry and resolves the scale uncertainty problem in single-view geometry. It is by combining two modalities that this method reaches semantic-geometric consistency, ensuring that the geometry as well as the semantics in the predicted voxels are correct \cite{xu2025survey, GenericADSurvey}.


However, because there is a considerable domain gap between the camera and LiDAR images, it has been challenging to effectively combine them. Camera systems produce dense data in a perspective image, while LiDAR sensors produce sparse data in free-space three-dimensional points. To combine them, it is necessary to express them in a common coordinate system. For image-based data, it involves a complex 2D-3D view transform, which has lately been facilitated using Lift-Splat-Shoot (LSS) \cite{kerbl20233dgaussians, kerbl20233dgaussians} or depth attention mechanisms \cite{miao2023occdepth}, where two-dimensional image features are first lifted to three-dimensional coordinates along viewing rays based on predicted depths. This directly fills a frustum volume, which is then splatted into a three-dimensional voxel or a Bird’s Eye View (BEV) \cite{huang2021bevdet, li2024bevformer} image. For LiDAR data, it is primarily processed by rasterizing an irregular and unordered set of points into regular data structures, such as voxels or vertical pillars with names sake pillars \cite{tang2024sparseocc}, allowing standard three-dimensional sparse convolution on points to extract geometric features, which are then aggregated into a common BEV \cite{huang2021bevdet, li2024bevformer} image with image data streams.


After having a shared representation, fusion approaches can be divided based on the point at which fusion happens. In early fusion, point-painting is done to directly project semantic labels or image feature representations onto raw 3D points before being fed to the network. Though seemingly a straightforward solution, early fusion approaches are highly vulnerable to synchronizing 3D point information in terms of their calibration and time. Moreover, early fusion unswervingly links the geometry pipeline to the potential errors in semantic image segmentation. Late fusion, or decision-level fusion, involves aggregating bounding box and/or occupancy grid outputs for independent sensor-specific backbones. In late fusion, outputs tend to be highly conflicting and difficult to combine using decision heuristics. Hence, in recent years, there has been a consensus in the research community to move to intermediate feature fusion for achieving better results. In BEVFuse \cite{li2024bevformer}, HyDRa \cite{unleashinghydra2024}, and similar approaches, deep feature representations learned through sensor-specific backbones are merged in a spatially aligned latent feature space to provide the network with information to learn texture-geometry complementarities even before coming up with a prediction.


Despite this degree of architectural maturity, however, there still persists a significant theoretical flaw with the prevailing structure of feature fusion models: the over-reliance on static fusion strategies. This is because all these models solely rely on fixed, deterministic fusion strategies like addition, concatenation, and static attention mapping that are indifferent to the quality of the incoming data. Such fixed nature violates the dynamism expected within the Operational Design Domain. For instance, where high rainfall is involved, the LiDAR point cloud is noisy due to the backscatter effect, while the camera images might still be reliable. However, where high glare is involved, photometric channels are degraded, but geometric ranging is reliable. An invariant fusion strategy that combines both channels equally will therefore yield noisy results. This points to an imminent but thus far unresolved need for the formulation of an uncertainty-aware fusion strategy that dynamically adjusts the fusion weights for the two sensors remotely while ensuring that the integrity of the 3D occupancy space estimation is maintained.

\section{The Representation Crisis and the Gaussian Paradigm}


In addition to the purely architectural aspects when dealing with sensor fusion, the question of what three-dimensional perception task can realistically achieve is inherently bound by the mathematical framework used to describe the environment. The presently available literature is organized in principle according to the use of a two-dimensional framework, specifically Voxels \cite{boeder2024gaussianflowocc}, Bird’s Eye View (BEV) maps \cite{huang2021bevdet, li2024bevformer}, and Tri-Perspective View (TPV) maps \cite{yuan2023tpvformer}. Voxel models are based on the idea of dividing the three-dimensional world into rigid cubes. While these preserve the three-dimensional structure, this framework is afflicted by the problem of dimensionality, requiring ($O(N^3)$) memory that is cubic in resolution, hence severely choosing between obtaining a coarse answer in a timely fashion or being suitable for real-time safety by being sufficiently low-latency. In order to reduce this overhead, most frameworks reduce the three-dimensional world to a two-dimensional one (BEV) \cite{huang2021bevdet}. This lossy compression elevates the problem in that the framework is no longer remotely suitable for describing environments involving complex vertical structures such as tunnel systems and vegetation overheads.


To mitigate the size limitation that comes with Bird’s-Eye-View (BEV) \cite{huang2021bevdet, li2024bevformer} projections while avoiding the $O(N^3)$ overhead that comes with voxel-based models, the Tri-Perspective View (TPV) \cite{yuan2023tpvformer} concept was formulated. TPV \cite{yuan2023tpvformer} involves casting projections from the three-dimensional environment along three orthogonal planes—top view, front view, and side view. This results in lower computational overhead, dropping from $O(N^3)$ complexity to $O(N^2)$ complexity. However, TPV \cite{yuan2023tpvformer} is essentially a discretized, axial approximation; for higher detail beyond the axes, feature points must be aggregated across the planes of projection, resulting in ghosting effects for complex environments.


In order to cope with the trade-off between resolution and efficiency, a novel paradigm \cite{kerbl20233dgaussians} has been developed in the form of 3D Gaussian Splatting. Unlike their voxel counterparts, which discretize the empty space, Gaussian Splatting has a Lagrangian (or particle) approach to this problem. The scene is modeled as a set of anisotropic three-dimensional Gaussians, wherein every primitive has a set of parameters to be learned specifically: a geometric center represented by ($\mu$), a covariance matrix ($\Sigma$), defining the three-dimensional geometry in space, an opacities scalar ($\alpha$), and a set of Spherical Harmonics, describing view-dependent colors.


Such a representation is implemented through designs like GaussianFormer \cite{huang2024gaussianformer}, which swaps the traditional fixed grid scanning pattern with a query-centric process. Rather than processing blank voxels in the millions, the network leverages a set of latent queries learnable by the AI. By using a series of cross-attention layers, the latent queries aggregate data from the sensor channels and make predictions about the parameters of the Gaussians within salient points in the scene \cite{gaussianformer3d2025}. The process achieves a decoupling of the memory requirements from the resolution of the scene, where processing power and the number of Gaussians in complex areas (e.g., a car) increase over simpler areas (e.g., a wall).


There are compelling reasons to incorporate a Gaussian representation within a multi-modal framework. Three-dimensional Gaussians serve as a best Common Basis for the integration of various data modalities. LiDAR provides the accurate geometric initializations required to ground the geometric variables of the Gaussians (parameters ($\mu$ and $\Sigma$)) in reality, thereby alleviating issues of floating artifacts that are normally present in vision-only methods. At the same time, image features are projected to enhance semantic properties and opacity. Such a harmonious combination of image and LiDAR information allows the framework to capitalize on the geometric reality and semantic expressiveness of both modalities within a single, differentiable framework that enables real-time visualization and occupation monitoring.

\section{Problem Statement and Thesis Contributions}


Although there has been considerable progress in multi-modal fusion and robust scene representation, there has been a missing link in their intersection, which is crucial for occupancy prediction tasks. This is because conventional occupancy prediction models generally suffer from one of two common structural issues in their architecture. First, although current state-of-the-art models (such as BEVFusion \cite{li2024bevformer}) incorporate computationally intensive voxel representations, they all use static fusion strategies, which makes them highly vulnerable in dynamic environmental settings, such as in rain or under light glares, where sensors tend to be unreliable due to variations in environmental factors. However, since environmental factors change dynamically, it is highly challenging for models to be robust in all such scenarios. Second, although novel models (GaussianFormer \cite{huang2024gaussianformer}) are based on Gaussians, they are generally designed for camera-centric configurations only. These models, being highly computationally efficient, tend to overlook the useful geometrical prior information encoded in LiDAR data, which, although robust, generally lack adaptively learnable techniques for effectively fusing sparse LiDAR depth maps with texture information. As a result, currently, none of the models offer a unified solution for effective occupancy prediction that has been enabled by attaining both the high efficiency of 3D Gaussians and robustness from adaptively learned.

\section{Proposed Methodology}\label{sec:proposed_method}


To help fill the gap between efficient 3D representation and robust multimodal perception, this dissertation presents a Gaussian-Based Adaptive Camera-LiDAR Multimodal 3D Occupancy Prediction method. The basic idea here is to fully take advantage of the complementary nature of the two different types of sensing modalities, which possess unique strengths in terms of semantic information for cameras versus geometric accuracy for LiDAR sensors, with a flexible representation that sidesteps the significant memory and computation required by dense voxel representations. By transforming both representations into a shared Gaussian space, this work offers advantages in terms of a representation that is both spatially adaptive (the Gaussians can expand and shrink dynamically depending on the level of complexity in the space) and uncertain in both shape and location. Essentially, this work has four tightly coupled components that seek to offset the difference in the sensors.

\begin{enumerate}





\item \textbf{LiDAR Depth Feature Aggregation:}A module using Depth-Wise Deformable Sampling and Stochastic Depth Partitioning to lift sparse geometric signals into Gaussian primitives.

\item \textbf{Entropy-Based Feature Smoothing:} This is a stochastic regularization technique that calculates cross-entropy in both directions to correct artifacts in a specific domain before fusion is performed.

\item \textbf{Adaptive Camera-LiDAR Fusion:} A multi-stage fusion module that uses dual-stream cross-attention for refinement and consistency-aware reweighting for handling conflicts between sensors.

\item \textbf{The Gauss-Mamba Head:} A new decoding stage using Tri-Perspective View decomposition and Selective State Space Models, also referred to as Mamba, to compensate for geometrical alignment errors.

\end{enumerate}

\subsection{LiDAR Depth Feature Aggregation}\label{subsec:depth_aggregation}


Generally, the frameworks based on Gaussians are more optimized for the dense rays of cameras, as the features are smoothly interpolated. The LiDAR point cloud is naturally sparse and irregular. In general, the application of the standard view aggregation strategy on the sparse LiDAR point cloud often leads to the loss of depth or the loss of the finer structure details. In this context, the LiDAR Depth Feature Aggregation (LDFA) module is proposed for this purpose. Unlike the traditional projection strategy, the proposed LDFA module utilizes the depth-wise deformable sampling strategy, where the features are sampled with offsets learned for the entire depth range of a voxel column. Stochastic Depth Partitioning is also utilized for the reduction of the inherent large noise-to-signal ratio in the sparse volumes. The features are aggregated through the Gated Global Fusion mechanism to bring the geometric signal values on top of the Gaussians, suppressing the empty space.

\subsection{Entropy-Based Feature Smoothing}\label{subsec:entropy_smoothing}

One of the key issues in multi-modal fusion is related to handling conflicting features, for example, a darker area in the image (shadow) which is geometrically ambiguous in LiDAR data. These conflicts often cause negative transfer problems, usually observed in models learned on noisy modalities generalizing to cleaner modalities. In such a problem, to better address the issue, some domain adaptation techniques~\cite{doruk2024transadapter, doruk2025mamba} and principles have helped in developing a novel concept named Entropy-Based Feature Smoothing in the Entropy-Based Feature Smoothing module. In this concept, bidirectional cross-entropy mapping functions ($\mathcal{H}_{cam \rightarrow lidar}$ and $\mathcal{H}_{lidar \rightarrow cam}$) have helped in determining feature disagreement between modalities. Higher entropy messages have helped in encouraging uncertainty in models. By fully using these two mapping functions, confidence weighting for feature channels has helped in handling feature streams in a connection layer based on a learnable scaling factor denoted as $\epsilon$. In this concept for greater robustness, a stochastic layer selection mechanism has helped in randomly applying layer-smoothing during training to avoid overfitting to fixed noise distributions.

\subsection{Adaptive Camera-LiDAR Fusion}\label{subsec:adaptive_fusion}


The previous modules are modality enhancement, but the combination of these requires handling heteroscedastic sensor noise. The reliability of sensors differs according to environmental factors, such that, for example, cameras work worse under night conditions, while LiDAR sensors work worse under heavy rain. Static concatenation is inappropriate for handling this. Therefore, this problem is solved with the proposed ACLF, which is a three-step process. First, the Dual-Stream Cross-Attention process allows geometric features to ask for texture, and vice versa, for improvement before merging. Second, the Soft Gating process produces a fusion map that dynamically changes between sensors according to environment reliability. Third, for eliminating ghosting issues like reflection, Consistency-Aware Reweight is applied, wherein the cosine similarity is used for eliminating the features that are actively different for the sensors.

\subsection{The Gauss-Mamba Head}
\label{subsec:gauss_mamba}


The final stage of the pipeline engages in the decoding of the dense 3D semantic grid of the occupancy. While the Transformer Head has the advantage of representing long dependencies accurately, it has the same limitations of high complexity ($O(N^2)$) in computation. In overcoming the high complexity of the Transformer Head, the Gauss-Mamba Head uses the linear complexity ($O(N)$) of the Selective State Space Models, described by \cite{gu2023mamba}. However, the anisotropic property of the geometric structure of driving environments requires the use of the Tri-Perspective View (TPV) Decomposition \cite{yuan2023tpvformer}, which decomposes the sparse Gaussian points into the three orthogonal plane channels of ($\mathcal{P}_{xy}, \mathcal{P}_{xz}, \mathcal{P}_{yz}$). On each of the channels, the use of the Raster Scanning approach encodes the unordered points into a serial form that goes through the Hierarchical Sparse Mamba U-Net to encode the global information. Finally, the Geometric Consensus Fusion combines the predictions of the overlapping perspectives by stabilizing the 3D points before projecting on the dense grid.

\section{Summary of Contributions}\label{sec:summary_contributions}
The primary contributions of this thesis are summarized as follows:

\begin{itemize}
\item \textbf{Gaussian-Based Unified Framework.} We present a camera--LiDAR 3D semantic occupancy prediction framework that operates natively in a Gaussian representation space, leveraging semantic features from cameras and geometric features from LiDAR in a unified, memory-efficient 3D abstraction.

\item \textbf{LiDAR Depth Feature Aggregation (LDFA).} We propose a module utilizing Depth-Wise Deformable Sampling and Stochastic Depth Partitioning. This design effectively resolves scale ambiguities and suppresses volumetric noise in sparse geometric data by learning content-based depth relationships.

\item \textbf{Entropy-Based Feature Smoothing.} We integrate a stochastic feature smoothing module that utilizes bidirectional cross-entropy to identify and mitigate domain-specific noise. This prevents negative transfer between conflicting modalities via a learnable residual correction.

\item \textbf{Adaptive Camera--LiDAR Fusion.} We design a comprehensive fusion architecture featuring Dual-Stream Cross-Attention for mutual refinement and Consistency-Aware Reweighting for conflict resolution, ensuring the model prioritizes reliable sensor data in dynamic environments.

\item \textbf{The Gauss-Mamba Architecture.} We introduce the first application of Mamba State Space Models~\cite{gu2023mamba} to Gaussian descriptors. Our head employs Tri-Perspective View (TPV) decomposition~\cite{yuan2023tpvformer} and Geometric Consensus to model global context with linear complexity, offering a scalable alternative to Transformer-based occupancy prediction.

\item \textbf{Empirical Validation and Benchmarking.} We demonstrate the state-of-the-art performance of our framework across the nuScenes~\cite{nuscenes} (OpenOccupancy~\cite{openoccupancy2023}, Occ3D~\cite{occ3d}) and SemanticKITTI~\cite{semantickitti} datasets, achieving a peak mIoU of 49.4\% on Occ3D. Our results show that the Gaussian-based representation is particularly effective at capturing irregular and thin structures—such as Vegetation (+8.2\% over OccFusion~\cite{occfusionming2024}) and Motorcycles—while maintaining superior robustness in safety-critical scenarios like rain and extreme darkness, where it outperforms multi-modal baselines by suppressing sensor-specific noise.
\end{itemize}

\section{Chapter Organization}


The following sections of this chapter will be organized such that a thorough basis for the proposed work is accomplished. The context for the newly emerging trend from sparse object detection towards dense 3D semantic occupancy prediction is provided in Section 1.1. Theoretical bounds of the dominant sensing modality, camera sensing, relative to LiDAR sensing, are described and analyzed in Section 1.2. Current multi-modal fusion techniques are surveyed, discussing the current reliance upon static fusion schemes, in Section 1.3. The problem of representation complexity for 3D perception is covered, with a brief introduction of 3D Gaussian Splatting, a continuous, computationally optimal representation, in Section 1.4. Finally, the formal problem definition is given within Section 1.5, and the proposed methodology is described fully in Section 1.6, before this chapter is summarized with a synthesized overview of all key thesis results in Section 1.7.
\chapter{Related Work}


The present chapter provides an extensive survey on the dynamic state-of-the-art in semantic 3D occupancy prediction, being divided into four major paradigms for research focus. Starting out, we review the contributions in Camera-Based solutions in terms of view transformation, depth-guided voxelization, and temporal memory in an attempt to distill dense geometry from RGB images, which are rich in semantic values yet depth-uncertain. Next, we review LiDAR-Based solutions in terms of exploiting strong geometry cues and sparse convolutional representations in order to capture complete structure independently from the intrinsically sparse point cloud data. Moving along, we review Multi-Modal fusion schemes in an attempt to incorporate the complementary benefits offered by camera and LiDAR modalities in a manner that efficiently deals with significant alignment and computation costs. Finally, we discuss the novel area of Gaussian Splatting, reviewing its benefits in being a differentiable, memory-efficient, and continuous alternative to voxel representations for complete structure capture. By integrating these advances, this chapter recognizes the drawbacks based on structural disparities and sensor differences to satisfactorily justify the common thread introduced in this thesis based on the use of Gaussians in a multi-modal framework.

\section{Camera Based 3D Occupancy Prediction}

\begin{figure}[ht]
    \centering
    \includegraphics[width=0.9\textwidth]{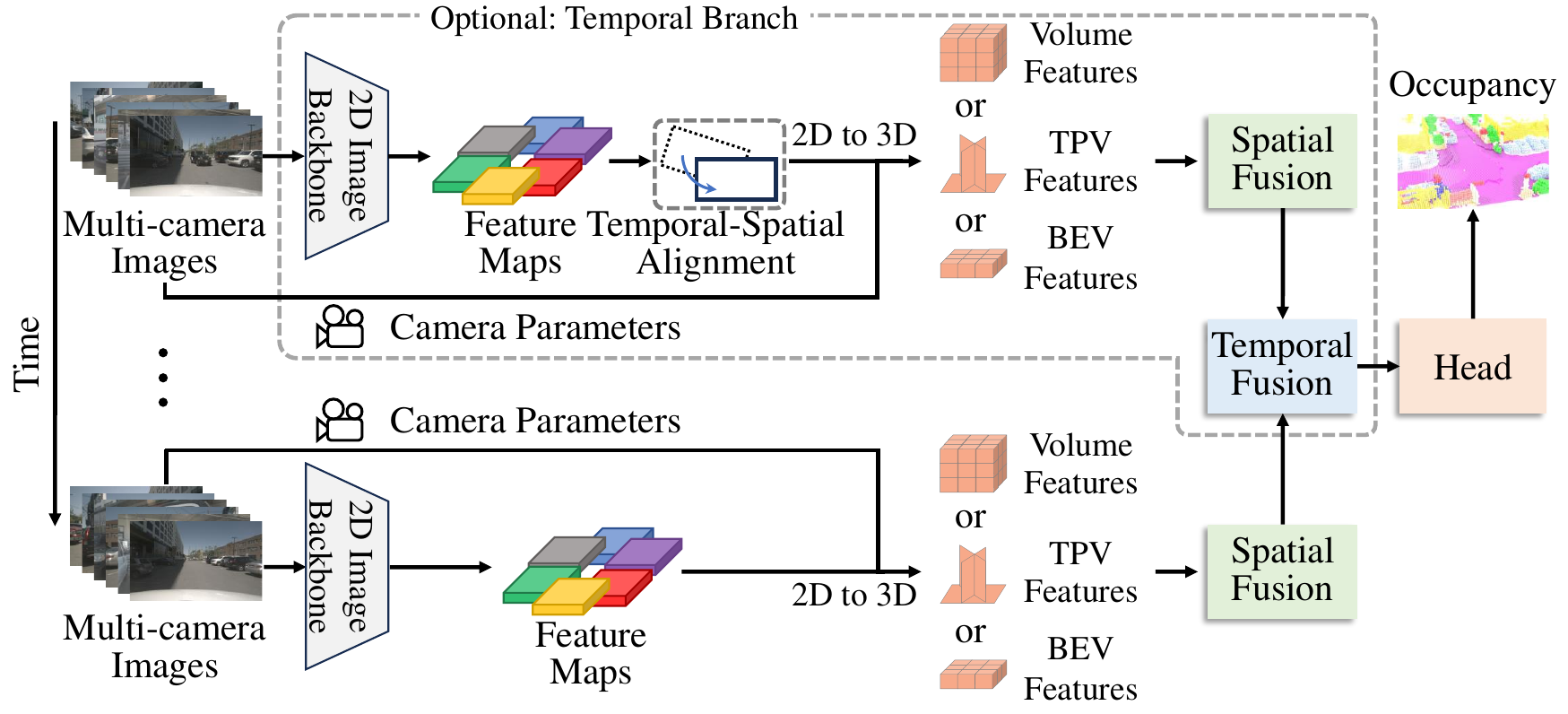}
    \label{fig:vision}
\end{figure}

Camera-based 3D semantic occupancy prediction has recently been identified as a competitive approach to traditional LiDAR-based solutions, given the economic viability and availability of vision sensors. This particular task aims to obtain voxel semantic information based on monocular or multi-view images.

Early success, represented by FB-OCC~\cite{li2023fbocc}, utilized forward \& backward view transforms to address the problem of occlusion \& depth cues in single-pass inference tasks, which improved the resulting geometric consistency in predicted outputs. OccDepth~\cite{miao2023occdepth} built upon this concept \& added depth estimation to guide voxel assignments to achieve semantic scene completion based solely on camera feeds.

In addition, to enhance the robustness of the model over time and make it easy to share the same data, the LMPOcc~\cite{yuan2025lmpocc} model included a new long-term memory prior, where the outputs of the perception modules for the past traversal are aggregated. The approach helps the model make decisions under challenging illumination conditions based on semantically aligned priors for past observations, as opposed to the short-term fusion approach of the traditional models.

Another important aspect is the spatial efficiency. SHTOcc~\cite{yu2025shtocc} proposed a sparse head-tail voxel representation, where computation is focused where the pixel is more uncertain or semantically significant. It is a sparse representation that removes redundant information, which is appropriate for real-time execution requirements.

There has also been an increasing trend in the adoption of generative representations. GaussRender~\cite{chambon2024gaussrender} and Gaussian Former-2~\cite{huang2024gaussianformer2} proposed the use of Gaussian representations for volumetric encoding and rendering, allowing for more flexible modeling of spatial distributions compared to the fixed voxel representation. This helps improve the representation of challenging regions, enabling better semantic completion in a probabilistic framework.

Resolution adaptability is another area where innovation has taken place. AdaOcc~\cite{chen2024adaocc} introduced adaptive resolution voxelization, in which varying density and resolution for feature representation across a scene allowed for efficient handling of large scenes while maintaining highly detailed areas around important objects and infrastructure.

Within these lines, one major disadvantage of camera-based approaches has persisted, namely the vulnerability to inaccuracies associated with monocular or stereo depth prediction. Further, the lack of high-fidelity geometric information, especially as it pertains to LiDAR-based alternatives, limits the geometric accuracy associated with occupancy prediction. This has contributed to decreased performance within complex geometries as well as conditions associated with sub-optimal visual quality.

\section{LiDAR Based 3D Occupancy Prediction}

\begin{figure}[ht]
    \centering
    \includegraphics[width=0.9\textwidth]{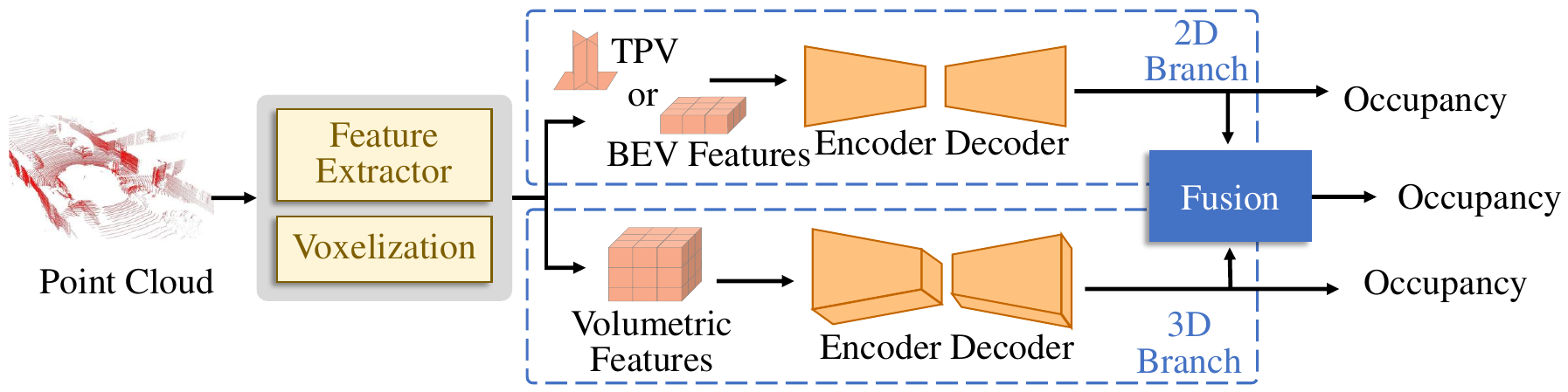}
    \label{fig:lidar}
\end{figure}

LiDAR-based 3D occupancy prediction has gained substantial momentum in autonomous driving because it accurately models geometric information. Earlier works, such as S3CNet~\cite{cheng2021s3cnet} and LMSCNet~\cite{roldao2020lmscnet}, focused on semantic scene completion using 3D convolution in voxels. These models learned to reconstruct full 3D semantic contents from sparse LiDAR data, using multiscale and sparse convolutional techniques to overcome the computational complexity required for dense voxel processing. Rist et al. \cite{rist2020deepimplicit} proposed models based on implicit functions, allowing for high-resolution recovery through prior shape models around local points.

Upon the success of single-frame inference, the integration of learned contextual shape priors for completion tasks into single-sweep segmentation models by Xu et al. \cite{xu2025survey} showcased the utility of geometric prior information for other perceptual tasks. However, representation learning soon matured to focus on more memory-efficient and compact representations rather than the dense voxel space. PointOcc~\cite{zuo2023pointocc} introduced a cylindrical tri-perspective view, which was specifically designed to suit the radial LiDAR scanning distribution, thereby assisting with point-level occupancy estimation. Around the same time, semantic and instance-level scene completion with the inclusion of uncertainty was provided by the solution of PaSCo~\cite{cao2023pasco}.

There has been a rising trend of models involving temporal reasoning and self-supervision. Occupancy-MAE~\cite{min2022occupancymae} proposed the masked autoencoding paradigm for pretraining on large-scale LiDAR data based on self-supervised learning objectives, resulting in better representation learning when observing sparsely. MetaSSC~\cite{qu2024metassc} further incorporated the meta-learning approach for adapting quickly to unseen environments with long sequence modeling of the temporal domain.

In addressing the problem concerning generalization and domain shifts, MergeOcc~\cite{xu2024mergeocc} conceptualized cross-device adaptation to fill the gap between dissimilar LiDAR sensors. This is imperative for deployment. Diffusion-based generative models also emerged. DiffSSC~\cite{cao2024diffssc} and DiffDistill~\cite{zhao2025diffdistill} apply probabilistic modeling that determines likely reconstructions within occluded regions by outperforming deterministic approaches since they can handle uncertainty.

Efficiency-focused models have also been introduced. OccRWKV~\cite{wang2024occrwkv} exploited a linear complexity recurrent attention approach utilizing RWKV to efficiently process long sequences, avoiding the quintic complexity penalty imposed by transformers. TALoS~\cite{jang2024talos} employed test-time adaptation utilizing geometrical constraints imposed by the line of sight in the LiDAR sensor. Even with such progress, LiDAR-based models for occupancy prediction remain prone to certain challenges. Since LiDAR is different from cameras, where rich semantic textures are captured, it is difficult for LiDAR sensors to discriminate between semantically different areas that appear similar (for example, roads and sidewalks). In addition, due to the natural sparsity of LiDAR point cloud data, the density of which is also range-dependent, there is quite a bit of performance degradation, especially for distant points or occluded areas.

\section{LiDAR-Camera Multi-Modal 3D Occupancy Prediction}

\begin{figure}[ht]
    \centering
    \includegraphics[width=\textwidth]{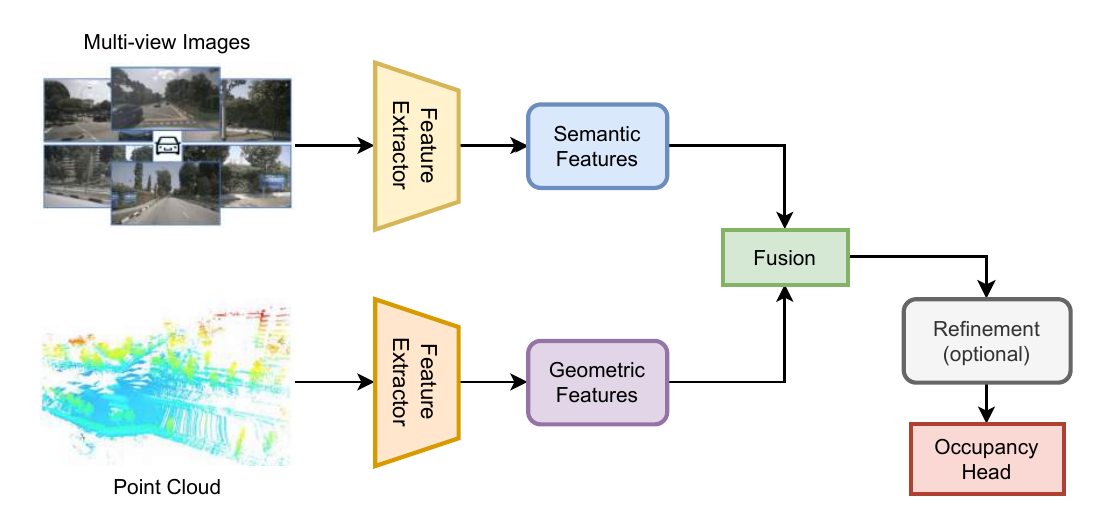}
\end{figure}

Multimodal semantic occupancy prediction in the 3D domain has thus turned out to be a crucial paradigm within the realm of autonomous driving and has played a critical role in gaining a complete and accurate understanding of the surroundings by making use of the complementarities existing among LiDAR, cameras, and radar. The initial work that brought into existence a large-scale benchmark and helped in making the evaluation process standardized among camera and LiDAR fusion-based models was that of OpenOccupancy~\cite{openoccupancy2023}.

In the year 2024, the community saw a general investigation of sensor fusion and architectural approaches. The HyDRa~\cite{unleashinghydra2024} system included radar as well as LiDAR and cameras, adding support for hybrid fusion and modules for depth consistency, proving robustness in challenging settings. The OccFusion~\cite{occfusionming2024} primarily utilized the extraction of sensor-specific features for fusion, while another relevant study of Zhang et al.~\cite{occfusionzhang2024} brought forward a depth estimation-free approach for more simplified fusion with less dependence on intermediate supervision. In addition to the above-said fusion-focused works, another relevant contribution of the year, the EFFOcc~\cite{effocc2024} system, concentrated on enhancing the use of weak labels for improved feasibility of deployment.

The need for efficiency and real-time processing was further highlighted by Sze et al.~\cite{realtimeocc2024}, who proposed a sparse convolutional network for real-time semantic occupancy estimation. LiCROcc~\cite{licrocc2024} demonstrated the applicability of LiDAR supervision for radar, improving occupancy estimation, while OccMamba~\cite{occmamba2024} brought a state-space model architecture, which allowed for the replacement of some attention-related components in favor of more efficient ones. DAOcc~\cite{daocc2024} presented a joint detection-occupancy architecture, where object detection information informed dense occupancy predictions, thus exploiting multitask learning power.

The fusion approach itself was also featured prominently as OccLoff~\cite{occloff2024} sought to optimize the weights of sensor-specific fusion, and Co-Occ~\cite{coocc2024} used volume rendering regularization. RobustOcc~\cite{robustocc2024} tackled issues related to calibration via transformation modules in space. Finally, PVP~\cite{pvp2024} made use of a polar coordinate framework that structured the input space in support of semantic understanding.

However, generative models also started to take off. OccGen~\cite{occgen2024} used a variational approach to represent semantic uncertainty, and MR-Occ~\cite{mrocc2024} introduced a hierarchical representation in terms of a voxel structure for scalable fusion. Later contributions started to make use of a wider range of sensors and tasks. Doracamom~\cite{doracamom2025} integrated multi-view radar and optical camera inputs to achieve omnidirectional perception, while AutoOcc~\cite{autoocc2025} used vision and language priors to tackle open-set semantic annotation through Gaussian splatting.

\sloppy
Meta-learning methodologies were brought forth with MetaOcc~\cite{metaocc2025}, with dual-training processes used for improved generalization over sensor modalities. OccCylindrical~\cite{occcylindrical2025} used cylindrical projection for preserving spatial information throughout fusion, while MinkOcc~\cite{minkocc2025} applied real-time sparse convolution for the first time for label-efficient scenarios. For the fusion of both geometric understanding and attention mechanisms, GaussianFormer3D~\cite{gaussianformer3d2025} combined deformable 3D attention with Gaussian features for adaptive modeling of complex contexts. Rounding off these milestones was OccLE~\cite{occle2025}, with semi-supervised training for efficient annotation. Notwithstanding these progresses, the task of multimodal occupancy prediction remains to be faced by several challenges. The LiDAR features tend to lack semantic depth because of the sparse point cloud representation, making it rather challenging to achieve a better understanding on the class level, especially at range or in the presence of severe occlusions. On the other hand, the camera features can potentially present semantic ambiguity along with uncertain depth in unstructured environments. Though multimodal fusion can help improve the robustness, it can bring a significant amount of added complexity to the computational overhead, especially when more complex encoder structures, multi-view warping, attention fusion modules, are used along with other scenarios. Another ultra-crisis point is going to be the domain gap problem between real-world LiDAR point cloud data and the semi-synthetic pseudo-LiDAR representations derived from single-view or stereo images. These domain-specific differences in geometric representation, noise patterns, can potentially reduce the uniformity in consistency levels along with efficiency for fusion strategies being used in such scenarios. In addition to this, semantic mismatch in modality representation will tend to raise more challenging tasks related to fusion on several grounds since LiDAR features are represented using voxel grids~\cite{occle2025,roldao2020lmscnet,openoccupancy2023}, bird’s-eye view representations~\cite{huang2021bevdet,li2024bevformer}, whereas camera features can presumably make use of perspective-agnostic token representations termed as TPV~\cite{yuan2023tpvformer}, along with 3D Gaussian splat representations.

\section{Gaussian Splatting}

In recent years, the technique of Gaussian splatting has also been recognized as one effective way for 3D environment representation and semantic occupancy prediction. In contrast to traditional voxel representation methods that divide the environment space into cubic voxels with fixed spatial resolution, Gaussian splatting adopts a sparse and continuous distribution of anisotropic three-dimensional Gaussians. A three-dimensional Gaussian is described by parameters such as its spatial coordinates, covariance value, color information, transparency information, and semantic semantics. This helps model real-world environments better since they often have sparse and regular occupancy \cite{huang2024gaussianformer, huang2024gaussianformer2}.

The mainstream way of doing occupancy prediction traditionally involves dividing a 3D space in a regular voxel grid or point cloud, which can lead to high memory costs and computational complexity owing to a large number of empty areas. While bird’s-eye view representation and 2D-to-3D lifting methods can reduce such problems, in many cases, they cause geometrical artifacts and are additionally limited by a low vertical resolution. The main advantage of Gaussian splatting, on the contrary, is that it can automatically distribute representation resources, achieving high detail representation in meaningful areas while discarding empty or homogeneous regions \cite{boeder2024gaussianflowocc}.

One major benefit offered by the Gaussian splatting approach has to do with its differentiability and ability to be integrated with projective rendering pipelines. Through the projection of three-dimensional Gaussians on planar image representations using perspective or orthographic projection cameras, the learning procedure can be efficiently supervised using planar ground truths, such as semantic or depth information, without requiring three-dimensional supervision. For this reason, this method can be truly useful in weakly supervised or self-supervised learning scenarios, especially in environments in which per-voxel three-dimensional supervision would be both expensive and unfeasible \cite{chambon2024gaussrender,jiang2024gausstr}. In addition, the rendering performed using Gaussian splatting ensures the consistency across multiple projections, which helps to better improve the spatial consistency in occupancy representation, often faced by camera-centric approaches.

Nonetheless, recent breakthroughs have increased the scope for applying Gaussian splatting over temporal domains. This has made it possible for models to learn dynamic scenes based on motion vectors (flows) associated with each Gaussian, making it easier for four-dimensional occupancy forecasting \cite{boeder2024gaussianflowocc,zheng2024gaussianad}. Furthermore, new works have begun investigating the integration of Gaussian splatting with foundation vision-language models. This has made it possible for models to use open vocabulary semantic occupancy prediction based on the alignment of rendered Gaussian projections with high-level features from large-scale image-text pairs \cite{jiang2024gausstr}.

\sloppy
The Gaussian splatting approach also showed promise in multi-modal datasets. Hybrid pipes that combine LiDAR and cameras have utilized the voxel-to-Gaussian initialization strategy for the representation of geometric priors as well as the refinement of semantic attributes with the help of texture cues emanating from cameras. The idea of deformable attention for the optimization of the lifting of the 3D spaces further enhances the integration of heterogeneous modalities of data as opposed to the voxel fusion approach \cite{zhao2025gaussianformer3d}.

Some methodologies have further exploited the use of the Gaussian representation to facilitate pretraining. By incorporating geometric and textural data in a single setting, pretraining methods developed from the concept of the Gaussian representation have demonstrated improvements in detection, mapping, and occupation tasks and bettered the efficiency of computations compared to methods using NeRFs \cite{xu2024gaussianpretrain}.

Despite the potential, there are several challenges in the Gaussian splatting method. First, the lack of direct surface constraints can lead to over-smoothed or "floating" artifacts in the presence of occlusions or ambiguities. Second, the quality of splatted models is strongly tied to the quality of initializations, an issue for systems operating in the monocular setting, in which depth inference is inherently ambiguous. Third, while there is a benefit of reduced computations in the inference step for the Gaussian splatting approach, the training process can still be computationally expensive, especially with the need for accurate rendering and backpropagation of the gradients. In sum, the Gaussian Splatting approach provides a strong and complex framework for the semantic task of 3D occupancy prediction. This capacity to combine geometrical, semantic, and temporal data in a sparse and differentiable form makes the approach among the most prominent in the development of 3D perception technology.

\section{Summary and Analysis}

There are three major paradigms in the methodology that frame the scope of 3D occupancy prediction based on the input modality and the techniques used. These are LiDAR-based approaches, camera-based approaches, and multi-modal fusion-based approaches. They have their own set of strengths and weaknesses.

RGB-based approaches often rely on the abundance and interpretability of camera images. Thus, they can be applied on a large scale because they do not require expensive hardware. However, recent improvements include better spatial and temporal consistency by means of depth information. Inadequate geometric information remains a limitation. Thus, accuracy regarding depth information and spatial detail can be problematic. It can be so especially when the environment is complex, far away, and less illuminated.

LiDAR-based methods provide a superior level of geometric accuracy and completeness because of the direct measurement of geometric cues. The LiDAR-based methods make use of the sparsity and radial sampling of point cloud data to build effective point-based/voxel-based models. The state-of-the-art developments have brought temporal models, self-supervised models, and generation-based models for better robustness of models when working with sparse point cloud data. The LiDAR-based point cloud models do not have dense semantic annotations, and the sparsity of LiDAR point cloud datasets, particularly for longer ranges, might hamper the interpretability of the LiDAR-based models when working with scenes with semantically ambiguous areas or similar semantic classes.

Multimodal approaches attempt to leverage for their benefit the semantic depth of camera observations and the geometric precision offered by LiDAR to overcome the limitations associated with unimodal approaches. Additionally, the fusion approaches make use of deep architectures tailored to both modalities along with innovative calibration and registration techniques to further boost robust occupancy prediction. Moreover, incorporating other modalities like radar and leverage provided by vision-and-language priors can further promote generalization. However, the process of modal fusion can give rise to challenges like calibration requirements, difficulties in harmonizing modal structures, such as registration between perspective view feature spaces and voxel grids. In conclusion, unimodal approaches are strong either in terms of semantic meaning or geometric information, yet the weakness in one modality hinders complete understanding in dynamic environments when dealing with 3D visual data. On the other hand, multimodal approaches provide a potential solution to this problem by combining the strengths from both paradigms, yet they pose a challenge in terms of managing their complexity in order to achieve real-time performance.
\chapter{Methodology}

\section{Framework Overview}

This thesis proposes a unified multi-modal perception framework designed to reconstruct dense 3D semantic occupancy from synchronized LiDAR point clouds and surround-view camera images. The primary challenge in multi-modal 3D perception is bridging the representational gap between dense, semantic-rich 2D images and sparse, geometrically accurate 3D point clouds. This discrepancy often leads to negative transfer, where noise or misalignment in one modality degrades the performance of the other.

To address these limitations, we introduce a method that eschews traditional fixed voxel queries in favor of a flexible Gaussian representation. Our approach integrates four novel contributions:
\begin{enumerate}
    \item \textbf{LiDAR Depth Feature Aggregation (LDFA):} A mechanism utilizing Depth-Wise Deformable Sampling and Stochastic Depth Partitioning to lift sparse geometric signals onto Gaussian primitives while suppressing volumetric noise.
    
    \item \textbf{Entropy-Based Feature Smoothing:} A stochastic regularization module that computes bidirectional cross-entropy maps to quantify inter-sensor uncertainty, dynamically rectifying domain-specific artifacts before fusion.
    
    \item \textbf{Adaptive Camera-LiDAR Fusion:} A multi-stage integration module employing Dual-Stream Cross-Attention for feature refinement and a Consistency-Aware Reweighting mechanism to filter sensor conflicts.
    
    \item \textbf{The Gauss-Mamba Head:} A geometric refinement head that leverages Tri-Perspective View (TPV) decomposition and Selective State Space Models (Mamba) to correct spatial misalignments with linear complexity.
\end{enumerate}

\begin{figure}[ht]
  \centering
  \includegraphics[width=\textwidth]{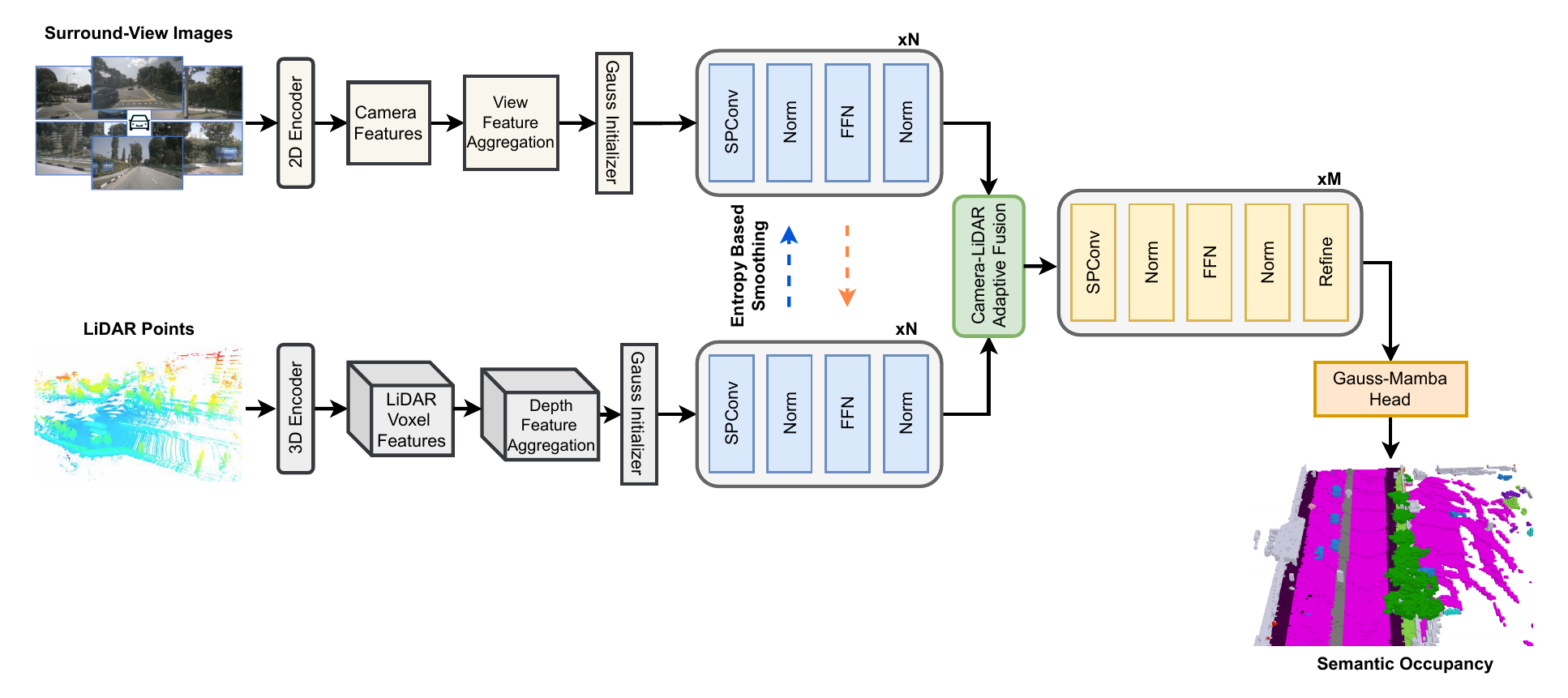}
\end{figure}

\subsection{Input Modalities}
The framework processes two distinct data streams, defined by specific coordinate systems and resolutions to ensure precise spatial alignment.

\subsubsection{Input Specifications}
\begin{itemize}
    \item \textbf{LiDAR Stream:} The geometric input consists of raw LiDAR point clouds clipped to a Cartesian range of $[-50.0m, 50.0m]$ along the $X$ and $Y$ axes, and $[-5.0m, 3.0m]$ along the $Z$ axis. To prepare this irregular data for processing, the continuous points are discretized into voxels with a resolution of $[0.1m, 0.1m, 0.1m]$.
    \item \textbf{Camera Stream:} The visual input comprises six synchronized surround-view images covering the full $360^{\circ}$ field of view. The images are pre-processed to an input resolution of $864 \times 1600$ to align with the network's stride requirements.
\end{itemize}

\subsection{Multi-Scale Feature Extraction}
The perception pipeline begins with the independent extraction of high-level feature representations from the two input modalities.

\textbf{Visual Stream:} To process the surround-view camera images, we employ a ResNet-101 backbone \cite{he2016deep} coupled with a Feature Pyramid Network (FPN) \cite{lin2017feature}. This results in robust, multi-scale embeddings that preserve both high-resolution textures and high-level semantic context.

\textbf{LiDAR Stream:} Simultaneously, the geometric stream utilizes a Hard Simple Voxel Feature Encoder (VFE) followed by a SparseEncoderHD \cite{graham20183d}. This encoder efficiently downsamples the grid while increasing the feature channel depth, extracting structural geometry without the computational waste of processing empty space.

\subsection{Gaussian Representation and Lifting}
A core innovation of our framework is the replacement of rigid voxel queries with sparse Gaussian primitives. These anchors are initialized at multiple resolutions and parameterized by position, scale, and semantic embeddings. To populate these anchors with environmental context, we employ two distinct branches:
\begin{itemize}
    \item \textbf{Camera Branch:} Multi-view appearance features are aggregated by projecting anchor centroids onto image planes, utilizing deformable attention to resolve perspective ambiguity.
    \sloppy
    \item \textbf{LiDAR Branch:} We introduce the LiDAR Depth Feature Aggregation (LDFA) module. This component employs depth-wise sampling to lift sparse geometric signals onto the Gaussian primitives while effectively filtering empty spatial volumes.
\end{itemize}

\subsection{Entropy-Based Feature Smoothing}
\sloppy
To address distributional misalignment between sensors—such as visual shadows that lack geometric counterparts—we introduce an Entropy-Based Feature Smoothing module. By calculating bidirectional uncertainty maps, the network identifies regions of sensor disagreement. These maps act as confidence weights to adaptively rectify feature streams, while a stochastic execution strategy provides regularization against over-fitting to modal artifacts.

\subsection{Adaptive Camera-LiDAR Fusion}
Rather than employing rigid concatenation, the Adaptive Camera-LiDAR Fusion Module treats sensor integration as a dynamic, context-aware process. This is achieved through:
\begin{enumerate}
    \item Dual-Stream Cross-Attention: Enabling mutual refinement where geometric features query texture information and vice-versa.
    \item Soft Gating: A learnable mechanism that modulates sensor dominance based on local reliability (e.g., prioritizing LiDAR in low-visibility conditions).
    \item Consistency Filtering: A channel-wise reweighting strategy based on cosine similarity to suppress ghost artifacts caused by sensor conflict.
\end{enumerate}

\subsection{The Gauss-Mamba Head}
The final stage utilizes the Gauss-Mamba Head to refine the attributes of the Gaussian primitives. To exploit the anisotropic nature of driving scenes, we perform a Tri-Perspective Decomposition, projecting the sparse coordinates into three orthogonal planes ($\mathcal{P}_{xy}, \mathcal{P}_{xz}, \mathcal{P}_{yz}$).

\section{Multi-Modal Feature Aggregation Strategy}
Our architecture utilizes a bifurcated processing stream to leverage the complementary nature of dense photometric data from cameras and sparse geometric data from LiDAR. Recognizing that these modalities exhibit fundamentally different topological structures—specifically, that cameras offer multi-view perspective while LiDAR offers multi-depth geometry—we employ distinct aggregation strategies tailored to the specific topology of each input.

\begin{figure}[ht]
  \centering
  \includegraphics[width=1\linewidth]{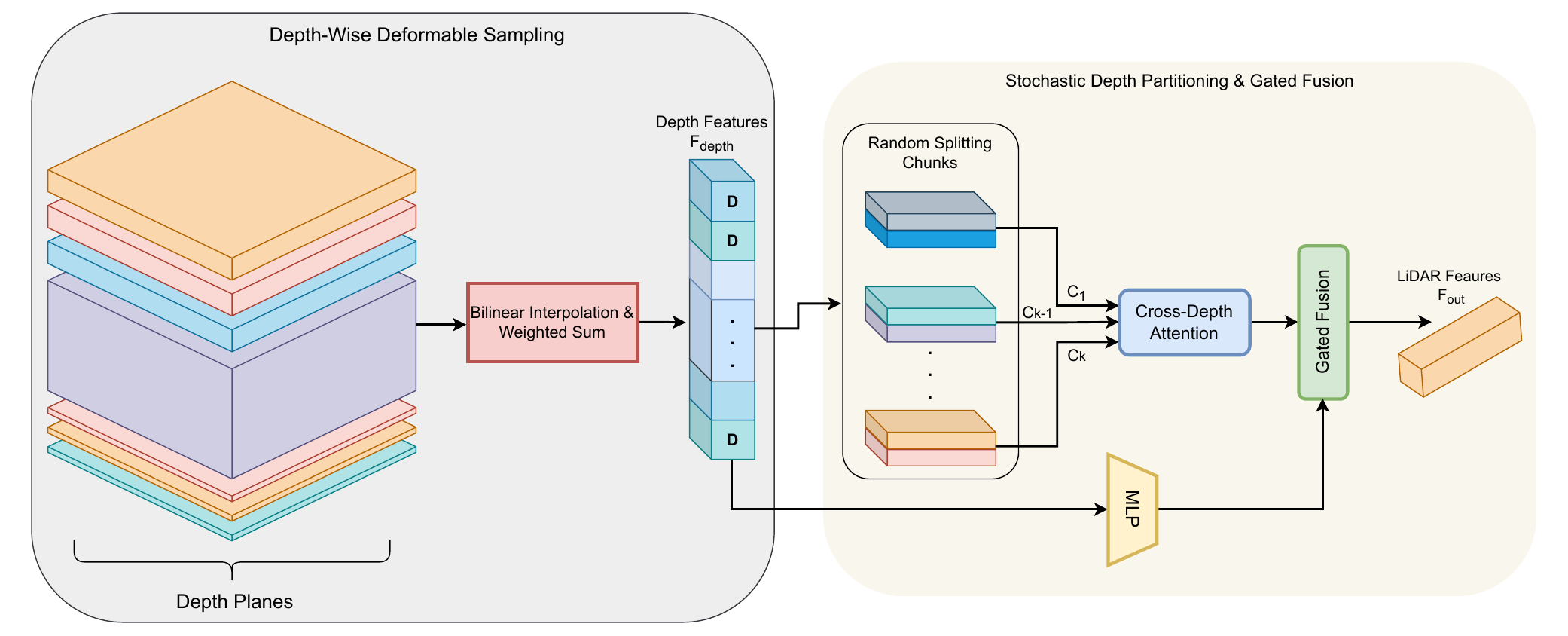}
\end{figure}

\subsection{Camera Branch: Multi-View Feature Aggregation}

For the image modality, which consists of $N=6$ surround-view images, we adopt the established Multi-View Feature Aggregation paradigm utilized in GaussianFormer \cite{huang2024gaussianformer}. Since 2D images lack explicit depth, the primary challenge in this domain is resolving perspective ambiguity, as a single 3D point often projects onto overlapping regions between adjacent cameras. This module treats 3D Gaussian anchors as spatial queries; by projecting the centroid of each anchor onto the intrinsic image planes, we extract and fuse appearance features using deformable attention, ensuring that the photometric attributes of the Gaussian primitives are consistent across all observing angles.

\subsection{LiDAR Branch: Depth Feature Aggregation}

In contrast to the camera branch, the LiDAR branch does not possess views. Instead, the raw point cloud is processed through voxelization and a 3D sparse encoder to generate a volumetric feature representation $\mathbf{V}$. This volume is strictly defined by the voxel grid size and the predetermined point cloud range. We interpret the vertical dimension of this volume as Depth, effectively creating a stack of $D$ distinct feature planes. Unlike camera images, where every pixel contains data, the LiDAR volume is inherently sparse; the majority of these depth planes correspond to empty space, while only a subset intersects with physical surfaces. To address this unique geometric distribution, we propose the LiDAR Depth Feature Aggregation (LDFA) module. This mechanism shifts the paradigm from aggregating views to aggregating depth strata, aiming to identify valid geometric signals across the voxelized depth intervals without losing fine-grained structural details.

To extract features from this volume, we employ a Depth-Wise Deformable Sampling strategy. For a given learnable Gaussian anchor $a_i$, we aim to extract features that describe its local geometry across the entire depth range. Rather than sampling a single point, a learnable offset generator predicts a set of $P$ local 3D keypoints around the anchor. We iterate through the $D$ depth levels, projecting these keypoints onto the corresponding feature plane of each level. Mathematically, let $\mathcal{P}_d(\mathbf{u}_{ik})$ denote the bilinear interpolation operation on depth plane $d$ at the projected keypoint coordinate $\mathbf{u}_{ik}$. The feature representation $\mathbf{f}_{i,d}$ for anchor $i$ at depth level $d$ is computed as:

\begin{equation}
    \mathbf{f}_{i,d} = \sum_{k=1}^{P} w_{ik} \cdot \mathcal{P}_d(\mathbf{u}_{ik})
\end{equation}

where $w_{ik}$ represents the learned attention weight for keypoint $k$. This operation results in a depth-stratified feature tensor $\mathbf{F}_{depth}$, which explicitly encodes the scene geometry as a sequence of $D$ depth hypotheses.

A significant challenge in processing this multi-depth tensor is the low signal-to-noise ratio caused by sparsity. A naive summation of the depth features would dilute the valid surface signal with noise from empty voxels. To resolve this, we utilize a Stochastic Depth Partitioning strategy. We partition the $D$ depth levels into $K$ distinct subgroups, or chunks, denoted as sets $\{S_1, \dots, S_K\}$. Crucially, to prevent the model from overfitting to fixed depth indices—for example, assuming the ground plane is always located at index 0—we apply a random permutation $\pi$ to the depth order during training. The representation for the $k$-th chunk, $\mathbf{C}_k$, is derived via mean pooling:

\begin{equation}
    \mathbf{C}_k = \frac{1}{|S_k|} \sum_{d \in S_k} \mathbf{F}_{depth}^{(\pi(d))}
\end{equation}

This strategy forces the network to learn relationships based on feature content rather than absolute position, significantly enhancing robustness.

To determine which chunk contains the actual surface geometry, we employ a Cross-Depth Attention mechanism. This functions as a content-based query system where specific depth chunks act as context to verify the existence of features in others. We project the initial chunks into a latent space to synthesize a Modulation Vector $\mathbf{M}$. This vector results from the interaction of different depth groups, conceptually defined as:

\begin{equation}
    \mathbf{M} = \phi(\mathbf{C}_1, \dots, \mathbf{C}_{K-1}) \odot \mathbf{C}_K
\end{equation}

where $\phi$ is a projection function and $\odot$ denotes element-wise multiplication. Theoretically, this operation amplifies feature magnitudes where the depth contexts agree, indicating a physical surface, and suppresses them where the chunks correspond to empty space. Finally, to ensure that the modulation process does not strictly remove potentially valuable information, we re-introduce the global context via a Gated Global Fusion mechanism. We compute a global mean feature $\mathbf{G}_{global}$ across all unperturbed depth levels. A learnable soft gate, $\alpha \in [0, 1]$, is derived via a sigmoid activation $\sigma$ to dynamically weigh the modulated features against the global mean:

\begin{equation}
    \mathbf{F}_{out} = \alpha \cdot \mathbf{M} + (1 - \alpha) \cdot \mathbf{G}_{global}
\end{equation}

This gated residual connection ensures that the model can focus on the specific depth containing the surface while retaining the structural context of the entire column, effectively lifting the sparse LiDAR geometry onto the 3D Gaussian primitives.

By synergizing the explicit geometric queries of the Depth-Wise Sampler with the noise-suppressing capabilities of the Chunk-Based Gated Fusion, the LDFA module provides a robust solution to the ill-posed problem of 2D-to-3D lifting in sparse geometric domains. Where standard methods often smear features along the projection ray due to depth ambiguity, our approach enforces a rigorous depth-consistency check. This ensures that the learned Gaussian primitives are initialized with features that are not only semantically rich but also geometrically grounded to the physical surfaces of the scene, thereby significantly improving the fidelity of the subsequent 3D occupancy prediction.

\section{Entropy-Based Feature Smoothing}

\begin{figure}[h]
  \centering
  \includegraphics[width=1\linewidth]{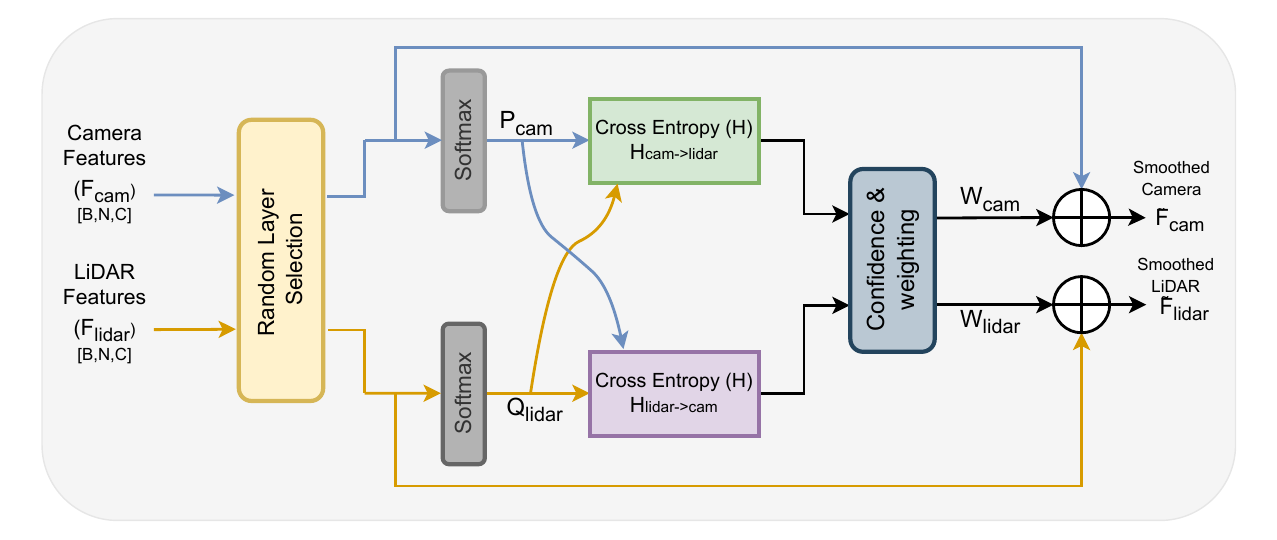}
\end{figure}

In multi-modal 3D perception, a fundamental challenge arises from the inherent heterogeneity of sensor data. Following the initial Gaussian Splatting stage, where Camera and LiDAR features are lifted into a shared 3D representation, the resulting feature maps often exhibit significant distributional misalignment. For example, a shadow might create a high-contrast visual edge in the camera stream that lacks any geometric correspondent in the LiDAR stream, or sparse LiDAR points may miss a semantic boundary clearly visible in an image. Standard fusion approaches often ignore these discrepancies, leading to negative transfer, where the noise or ambiguity in one modality corrupts the robust features of the other. To address this, we introduce the Entropy-Based Feature Smoothing module, which operates intermittently between the Gaussian Splatting and the final Fusion stages. Its primary objective is to quantify the uncertainty between the two modalities using a cross-entropy metric and dynamically rectify the feature space to suppress rigid, domain-specific artifacts before fusion.

To ensure the model learns robust representations and to prevent overfitting to specific feature artifacts, we implement a stochastic execution strategy. Drawing inspiration from the regularization mechanisms introduced in TransAdapter \cite{doruk2024transadapter}, we do not apply feature smoothing uniformly across all encoder layers. Instead, we employ a randomized selection process where, during each training iteration, a subset of layers is dynamically chosen to undergo smoothing while others are bypassed. This stochasticity is critical for two reasons: first, it prevents the network from becoming over-dependent on the smoothing module to correct feature misalignments, forcing the encoder to learn intrinsically robust features. Second, it acts as a regularization constraint that prevents the model from overfitting to the noise patterns of a specific layer, ensuring that the feature manipulation remains generalized and effective across varying scene complexities.

The core mechanism for quantifying the discrepancy between the Camera and LiDAR streams is Cross-Entropy (CE). In this context, we interpret the normalized feature vectors of each Gaussian anchor as probability distributions over the latent semantic space. High cross-entropy between these distributions serves as a proxy for high uncertainty or disagreement between the sensors. Let $\mathbf{F}_{cam}$ and $\mathbf{F}_{lidar}$ denote the feature sets. We first convert these raw logits into probability distributions $P_{cam}$ and $Q_{lidar}$ using the Softmax function along the channel dimension, scaled by a temperature parameter $\tau$ to control distributional sharpness:

\begin{equation}
    P_{cam} = \text{Softmax}\left(\frac{\mathbf{F}_{cam}}{\tau}\right), \quad Q_{lidar} = \text{Softmax}\left(\frac{\mathbf{F}_{lidar}}{\tau}\right)
\end{equation}

Based on these distributions, we calculate the bidirectional cross-entropy. This yields two distinct entropy maps, each offering specific insight into the sensor disagreement. The Camera-Dominant Entropy, $\mathcal{H}_{cam \rightarrow lidar}$, treats the camera distribution as the target and measures how well the LiDAR features align with the visual data. A high value here implies that the Camera detects a clear semantic pattern that the LiDAR features fail to support. Conversely, the LiDAR-Dominant Entropy, $\mathcal{H}_{lidar \rightarrow cam}$, uses the geometric features as the reference. A high value in this direction indicates that the geometric structure is confident, but the visual features are ambiguous, perhaps due to lighting conditions or glare:

\begin{equation}
    \mathcal{H}_{cam \rightarrow lidar} = - \sum_{c=1}^{C} P_{cam}^{(c)} \cdot \log\left(Q_{lidar}^{(c)} + \xi \right)
\end{equation}

\begin{equation}
    \mathcal{H}_{lidar \rightarrow cam} = - \sum_{c=1}^{C} Q_{lidar}^{(c)} \cdot \log\left(P_{cam}^{(c)} + \xi \right)
\end{equation}

These entropy maps are then transformed into confidence scores to guide the feature manipulation. We employ an exponential decay function, where high entropy results in low confidence, followed by a normalization step to derive the relative modulation weights $W_{cam}$ and $W_{lidar}$:

\begin{equation}
    \omega_{cam} = \exp(-\mathcal{H}_{lidar \rightarrow cam}), \quad \omega_{lidar} = \exp(-\mathcal{H}_{cam \rightarrow lidar})
\end{equation}

\begin{equation}
    W_{cam} = \frac{\omega_{cam}}{\omega_{cam} + \omega_{lidar} + \xi}, \quad W_{lidar} = \frac{\omega_{lidar}}{\omega_{cam} + \omega_{lidar} + \xi}
\end{equation}

Finally, the features are updated via a residual connection that integrates these confidence weights. Crucially, we introduce a learnable scaling parameter $\epsilon$ (epsilon) to control the magnitude of this update. The motivation for making $\epsilon$ learnable, rather than a fixed hyperparameter, is to allow the network to adaptively determine the optimal intensity of smoothing during backpropagation. This ensures that the module can gently refine features in early training stages and apply stronger corrections as the alignment improves, without overwhelming the original signal:

\begin{equation}
    \tilde{\mathbf{F}}_{cam} = \mathbf{F}_{cam} + \epsilon \cdot W_{cam}
\end{equation}

\begin{equation}
    \tilde{\mathbf{F}}_{lidar} = \mathbf{F}_{lidar} + \epsilon \cdot W_{lidar}
\end{equation}

\section{Adaptive Camera-LiDAR Fusion Module}

In the pursuit of robust 3D scene reconstruction, relying on a single sensing modality creates critical perception bottlenecks, particularly when autonomous systems operate in unstructured or adverse environments. LiDAR sensors serve as the gold standard for geometric acquisition, capturing precise metric depth and structural details; however, they inherently suffer from resolution sparsity at range and provide no semantic texture information. Conversely, RGB cameras offer high-resolution textural and semantic context but are fundamentally limited by scale ambiguity and are susceptible to severe degradation under challenging lighting conditions, rain, or fog.

While multi-modal fusion aims to synthesize the complementary strengths of these sensors, traditional approaches often employ rigid aggregation mechanisms—such as direct feature concatenation or element-wise summation. These methods operate on the fragile assumption that both sensors provide valid, high-quality information at every point in space. This assumption frequently fails in real-world scenarios: for instance, concatenating features from a rain-obscured camera lens with valid LiDAR geometry introduces destructive noise into the latent space, while rigid multiplicative fusion can lead to irreversible signal loss (the vanishing feature problem) if one modality fails to detect an object entirely.

To address these limitations, we propose the Adaptive Camera-LiDAR Fusion Module, a unified architecture designed to treat fusion as a dynamic, context-aware process rather than a static operation. Our approach is motivated by the insight that robust fusion requires not only the aggregation of features but also the explicit modeling of consistency between modalities to suppress artifacts when sensors disagree.

\begin{figure}[ht]
    \centering
    \includegraphics[width=1\textwidth]{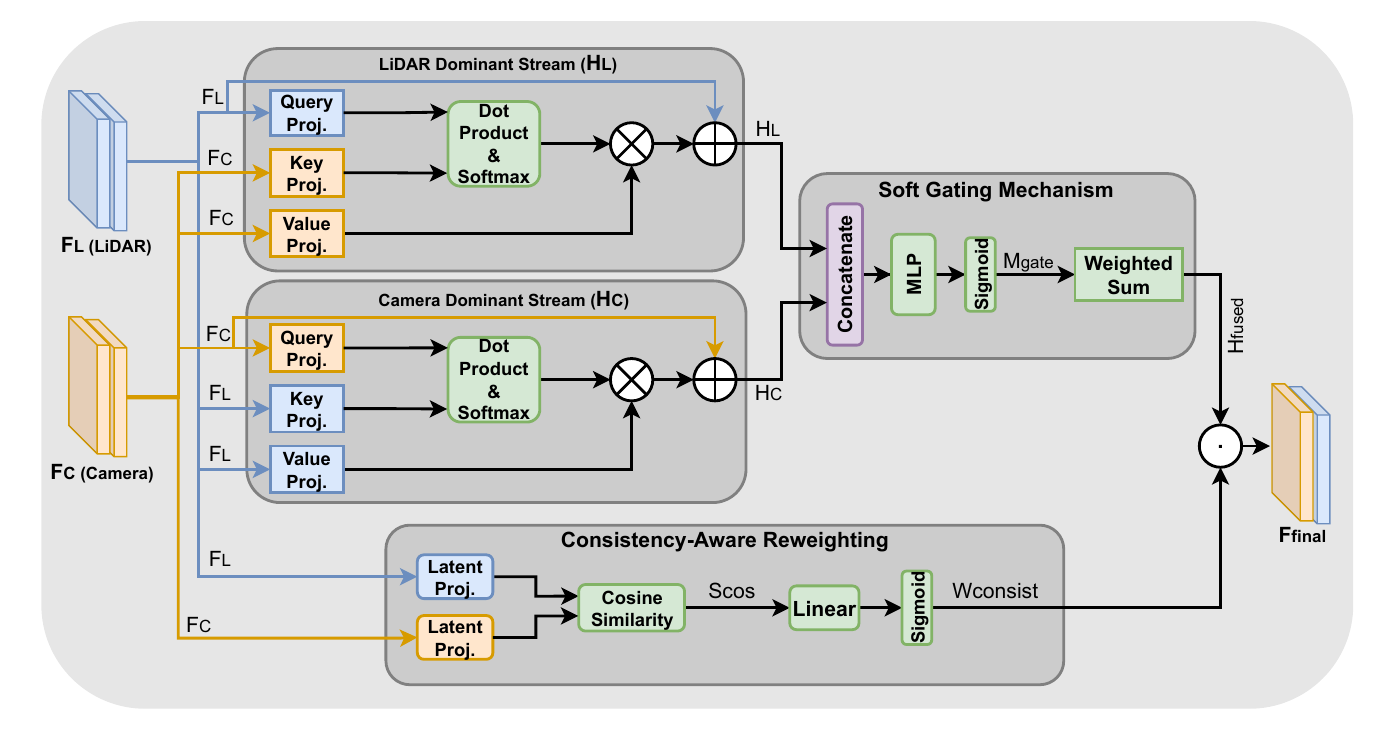}
\end{figure}

The first stage of our pipeline addresses the need for mutual feature enhancement through a Dual-Stream Cross-Attention mechanism. In standard fusion, geometric and textural features are often merged prematurely, forcing the network to resolve domain discrepancies immediately. Instead, we establish two parallel processing streams to refine each modality using context from the other before aggregation.

In the LiDAR-dominant stream, geometric features serve as queries ($Q$) seeking complementary texture information from the Camera (Values $V$). This formulation allows geometric points to acquire semantic color properties—such as determining that a flat surface is a ``road''—without corrupting their structural integrity. Simultaneously, in the Camera-dominant stream, visual features act as queries seeking LiDAR geometry to resolve depth ambiguities. Critically, unlike standard Vision Transformers which incur quadratic computational complexity ($O(N^2)$), we leverage the spatial alignment of our 3D anchors to implement an efficient point-wise attention mechanism ($O(N)$).

The refinement for the LiDAR-dominant stream ($H_L$) is defined as:

\begin{equation}
    Q_L = W_{q}^L F_L, \quad K_C = W_{k}^C F_C, \quad V_C = W_{v}^C F_C
\end{equation}

\begin{equation}
    H_L = F_L + \left( \sigma \left( \frac{Q_L \cdot K_C}{\sqrt{d}} \right) \odot V_C \right)
\end{equation}

Symmetrically, the Camera-dominant stream ($H_C$) extracts structural context from the LiDAR features:

\begin{equation}
    Q_C = W_{q}^C F_C, \quad K_L = W_{k}^L F_L, \quad V_L = W_{v}^L F_L
\end{equation}

\begin{equation}
    H_C = F_C + \left( \sigma \left( \frac{Q_C \cdot K_L}{\sqrt{d}} \right) \odot V_L \right)
\end{equation}

The use of residual connections in both equations is architecturally significant: it ensures that the fusion is additive rather than destructive. In scenarios where the cross-modal context is irrelevant or noisy (e.g., a camera looking into pitch darkness), the attention weights approach zero, preserving the original primary features ($F_L$ or $F_C$) and preventing the signal degradation common in multiplicative approaches.

Following this bi-directional refinement, the network must synthesize the two streams into a single coherent representation. To avoid the instability of rigid averaging, we employ a Soft Gating Mechanism. Rigid operations fail to account for the spatially varying reliability of sensors; for example, LiDAR is more reliable on textureless walls, while cameras are more reliable on distant traffic signs where LiDAR points are sparse. We address this by concatenating the refined streams and projecting them through a learned Multi-Layer Perceptron (MLP) to predict a scalar fusion mask $\mathcal{M}_{gate}$. This allows the network to learn a flexible policy, effectively switching dominance between sensors on a point-by-point basis:

\begin{equation}
    H_{fused} = \mathcal{M}_{gate} \odot H_L + (1 - \mathcal{M}_{gate}) \odot H_C
\end{equation}

Finally, we address the challenge of Sensor Conflict, a phenomenon where modalities actively contradict each other—such as a mirror reflection appearing as a valid object in the camera but as empty space in LiDAR. Standard attention mechanisms may inadvertently fuse these hallucinations, leading to ghost artifacts in the final occupancy grid. To mitigate this, we introduce a Consistency-Aware Reweighting prior. We calculate the cosine similarity between the latent projections of the input modalities to measure their semantic consensus. A high similarity score implies consensus, while a low score indicates a potential conflict or outlier.

Crucially, we do not use this raw score as a global volume knob, as that would uniformly suppress all information. Instead, we project the similarity score through a learnable linear layer to generate a Channel-Wise Consistency Gate, $\mathcal{W}_{consist}$. This enables the model to apply a nuanced filtering strategy: it can learn to be strict with geometric channels (suppressing them if LiDAR and Camera disagree on depth) while being lenient with semantic channels.

\begin{equation}
    F_{final} = H_{fused} \odot \mathcal{W}_{consist}
\end{equation}

This final reweighting acts as a learned noise-suppression filter, ensuring that the features propagated to the decoder are verified for multi-modal consistency, thereby improving the robustness of the final Gaussian representation.

\section{Gauss-Mamba Head}
\label{sec:head}


To address the challenge of high-fidelity 3D occupancy prediction, we propose a coarse-to-fine framework that leverages the explicit geometric nature of 3D Gaussian Splatting combined with the efficiency of State Space Models. Our pipeline begins with a set of fused, multi-modal features derived from the upstream perception backbone. To lift these features into a sparse 3D representation, we adopt the query-based generation head from GaussianFormer \cite{huang2024gaussianformer} as our baseline generator. We initialize a set of $N$ learnable 3D queries, denoted as $\mathcal{Q} \in \mathbb{R}^{N \times C}$, which interact with the fused feature volume via deformable cross-attention. This process yields a set of $N$ coarse Gaussian primitives, where each primitive $i$ is parameterized by a tuple $\mathbf{g}_i = \langle \boldsymbol{\mu}_i, \mathbf{s}_i, \mathbf{q}_i, \alpha_i, \mathbf{c}_i \rangle$. Here, $\boldsymbol{\mu}_i \in \mathbb{R}^3$ represents the centroid position, $\mathbf{s}_i$ the scaling factors, $\mathbf{q}_i$ the rotation quaternion, $\alpha_i$ the opacity logit, and $\mathbf{c}_i$ the semantic logits. While this baseline effectively identifies the general existence of objects, the geometric precision of the generated primitives often suffers from spatial jitter and lack of structural continuity due to the independent nature of the query processing mechanism. To rectify these geometric misalignments, we introduce our core contribution: the Gauss-Mamba  Head.

The design of our refinement head is architecturally grounded in the Tri-Perspective View (TPV) decomposition, originally proposed in TPVFormer  \cite{yuan2023tpvformer} for dense voxel grids. We adapt this philosophy to the domain of sparse, unstructured Lagrangian points to exploit the inherent geometric anisotropy of 3D driving scenes. In urban environments, semantic information is not distributed uniformly; the structural correlations on the horizontal ground plane (e.g., road topology) are fundamentally distinct from those on vertical planes (e.g., pole verticality). To decouple these conflicting geometric priors, we decompose the refinement problem into three orthogonal planar sub-branches ($\mathcal{P}_{xy}$, $\mathcal{P}_{xz}$, and $\mathcal{P}_{yz}$). For each primitive with centroid $\boldsymbol{\mu}_i = [x_i, y_i, z_i]^\top$, we project the continuous coordinates onto the planes via projection functions $\pi_p$, avoiding the quantization errors inherent in voxel-based methods:

\begin{equation}
\mathbf{v}_i^{xy} = [x_i, y_i]^\top, \quad \mathbf{v}_i^{xz} = [x_i, z_i]^\top, \quad \mathbf{v}_i^{yz} = [y_i, z_i]^\top
\end{equation}

We embed these projected coordinates into high-dimensional feature spaces using plane-specific Multi-Layer Perceptrons (MLP), creating three distinct feature manifolds $\mathbf{h}_i^p = \Phi_p(\mathbf{v}_i^p)$ that allow the model to learn plane-specific geometric contexts without the computational burden of processing empty space.

The most critical innovation of our framework is the integration of Dynamic TPV Scanning with the Selective State Space Model (Mamba). While Mamba offers linear computational complexity $O(N)$, enabling the processing of dense scenes with over 25,000 primitives, it inherently requires a causal 1D sequence. An unordered point cloud lacks this causal structure. To bridge this gap, we introduce a Raster Scanning strategy that transforms the set of sparse points into a spatially correlated sequence. For each plane, we serialize the primitives based on their spatial proximity; for instance, in the $\mathcal{P}_{xy}$ branch, we sort primitives primarily by their $Y$-coordinate and secondarily by their $X$-coordinate. Mathematically, the sorting indices $\mathcal{I}^{xy}$ are computed on detached coordinates using a stop-gradient operator $\text{sg}(\cdot)$ to ensure training stability:

\begin{equation}
\mathcal{I}^{xy} = \mathop{\mathrm{argsort}}_{i \in \{1 \dots N\}} \left( \text{sg}(y_i) \cdot \Omega + \text{sg}(x_i) \right)
\end{equation}

Once serialized, the sequence is processed by a Sparse Mamba U-Net. This architecture employs a symmetric Encoder-Bottleneck-Decoder design where the bottleneck utilizes the Mamba algorithm. Mamba discretizes the continuous state space equation $h'(t) = \mathbf{A}h(t) + \mathbf{B}x(t)$ using the Zero-Order Hold (ZOH) method, where the discretization parameters become functions of the input:

\begin{equation}
\overline{\mathbf{A}} = \exp(\Delta \mathbf{A}), \quad \overline{\mathbf{B}} = (\Delta \mathbf{A})^{-1}(\exp(\Delta \mathbf{A}) - \mathbf{I}) \cdot \Delta \mathbf{B}
\end{equation}

This mechanism allows the network to scan the scene, correcting geometric errors by understanding both local object shapes and global scene layout.

Following the independent refinement of features, we employ a Geometric Consensus Fusion strategy to update the 3D positions. Since each spatial axis is covered by two orthogonal planes (e.g., the $z$-axis is refined by both $\mathcal{P}_{xz}$ and $\mathcal{P}_{yz}$), we compute the final coordinate offsets as the average of the predictions from the contributing views. For a primitive $i$, the updated centroid $\boldsymbol{\mu}_i^{new}$ is derived via linear projection heads $\Psi$:

\begin{equation}
\boldsymbol{\mu}_i^{new} = \boldsymbol{\mu}_i + \frac{1}{2} \begin{bmatrix} 
\Psi_{x}^{xy}(\mathbf{h}_{out}^{xy}) + \Psi_{x}^{xz}(\mathbf{h}_{out}^{xz}) \\ 
\Psi_{y}^{xy}(\mathbf{h}_{out}^{xy}) + \Psi_{y}^{yz}(\mathbf{h}_{out}^{yz}) \\
\Psi_{z}^{xz}(\mathbf{h}_{out}^{xz}) + \Psi_{z}^{yz}(\mathbf{h}_{out}^{yz})
\end{bmatrix}
\end{equation}

This consensus mechanism acts as a geometric regularizer, ensuring structural consistency. Upon stabilizing the geometry, the refined features are projected into a 28-channel output vector for each primitive, encapsulating the centroid offsets, log-scales, rotation quaternions, opacity, and semantic logits. Finally, to bridge the continuous representation with the discrete ground truth for supervision, we employ a differentiable Gaussian-to-Voxelization process. The influence of the $i$-th Gaussian at a voxel coordinate $\mathbf{x}$ is given by the multivariate normal evaluation weighted by its local existence probability $p_i$:

\begin{equation}
G_i(\mathbf{x}) = p_i \cdot \exp \left( -\frac{1}{2} (\mathbf{x} - \boldsymbol{\mu}_i^{new})^\top \mathbf{\Sigma}_i^{-1} (\mathbf{x} - \boldsymbol{\mu}_i^{new}) \right)
\end{equation}

This final rasterization step aggregates the covariance, opacity, and semantic attributes of overlapping primitives to generate the final dense semantic occupancy volume.

\section{Objective Function}
\label{sec:loss}

To train the our framework effectively, we employ a composite objective function that simultaneously optimizes voxel-wise classification accuracy and the intersection-over-union (IoU) metric. Given the extreme class imbalance inherent in the 3D semantic occupancy task—where empty space vastly outweighs occupied voxels, and common objects like cars outnumber rare classes like bicycles—a single loss function is often insufficient. Therefore, our total loss $L_{total}$ is formulated as a weighted sum of the Weighted Cross-Entropy Loss ($L_{ce}$) and the Lovász-Softmax Loss ($L_{lovasz}$):

\begin{equation}
    L_{total} = \lambda_{ce} \cdot L_{ce} + \lambda_{lovasz} \cdot L_{lovasz}
\end{equation}

where $\lambda_{ce}$ and $\lambda_{lovasz}$ are hyperparameters controlling the contribution of each component. Based on our empirical settings, we assign $\lambda_{ce} = 10.0$ and $\lambda_{lovasz} = 1.0$ to balance the gradient magnitude between voxel-wise supervision and global metric optimization.

\subsection{Weighted Cross-Entropy Loss}
The primary supervision signal is provided by the Weighted Cross-Entropy loss. Standard Cross-Entropy treats all classes equally, which causes the model to bias towards the majority classes (e.g., drivable surface or background). To counteract this long-tail distribution problem, we incorporate explicit class weights $w_c$. The loss for a single voxel $i$ is defined as:

\begin{equation}
    L_{ce} = - \sum_{c=1}^{C} w_c \cdot y_{i,c} \log(p_{i,c})
\end{equation}

where:
\begin{itemize}
    \item $C$ is the total number of classes (18, including empty).
    \item $y_{i,c}$ is the binary ground truth indicator (1 if voxel $i$ belongs to class $c$, else 0).
    \item $p_{i,c}$ is the predicted probability for class $c$.
    \item $w_c$ is the manual balancing weight for class $c$. As defined in our configuration, rare classes such as construction vehicle ($w \approx 1.30$) and bicycle ($w \approx 1.27$) are assigned higher weights compared to common classes, enforcing the model to focus on learning difficult examples.
\end{itemize}

\subsection{Lovász-Softmax Loss}
While Cross-Entropy optimizes individual voxel accuracy, the evaluation metric for semantic occupancy is the mean Intersection-over-Union (mIoU). To bridge the gap between the optimization objective and the evaluation metric, we utilize the Lovász-Softmax Loss \cite{berman2018lovasz}. This loss serves as a differentiable surrogate for the Jaccard Index (IoU), allowing direct optimization of the metric during training.

\begin{equation}
    L_{lovasz} = \frac{1}{|C|} \sum_{c \in C} \overline{\Delta_{J_c}}(m(c))
\end{equation}

where $\overline{\Delta_{J_c}}$ represents the Lovász extension of the Jaccard loss for class $c$, and $m(c)$ is the vector of pixel errors. Crucially, we exclude the empty class (label 17) from the Lovász calculation, ensuring that the IoU optimization focuses purely on the geometric and semantic accuracy of occupied objects.

\chapter{Experiments}


In this chapter, a comprehensive empirical assessment of the proposed GaussianOcc3D framework on various large-scale autonomous driving benchmarks such as nuScenes~\cite{nuscenes} and SemanticKITTI~\cite{semantickitti} is provided. It starts with a description of the experimental settings, implementation details, and mean Intersection over Union (mIoU) metrics used to measure both semantic and geometric precision. However, evaluation centers on a comparison study against other existing camera-only, LiDAR-only, and multi-modal approaches to showcase the effectiveness in 3D reconstruction achieved through the Gaussian-based framework. In order to validate the chosen architectural details, extensive ablation studies have been carried out to demarcate the effect contributions of Adaptive Camera-LiDAR Fusion (ADA-Fusion), LiDAR Depth Feature Aggregation (LDFA), and Gauss-Mamba Head to the entire framework. Moreover, robustness analysis is discussed in terms of specialized performance analyses for safety-critical scenarios induced by rain and low-light environments. Lastly, prediction precision in terms of Gaussian primitive density and computational complexity analyses have been discussed to validate efficiency and feasibility for real-time autonomous perception applications.

\section{Datasets}


The proposed framework's capability of performing and generalizing well is examined on the basis of two large-scale benchmarks: nuScenes~\cite{nuscenes} and SemanticKITTI~\cite{semantickitti}. Notably, both of these datasets introduce different levels of complexities to the task of 3D occupancy prediction. For instance, the nuScenes dataset involves a multi-modal and multi-view camera and LiDAR setup that is exposed to varying weather and illumination conditions. In contrast to that, the SemanticKITTI dataset involves high-resolution LiDAR point cloud data and voxel annotations that are densely packed in complex European environments.

\subsection{NuScenes Dataset}


The dataset used for this study is called the NuScenes~\cite{nuscenes} dataset. This is another large-scale dataset for self-driving cars. It has recorded data in two different locations geographically: Boston, USA, and Singapore. The two locations have complementary attributes. While Boston has complex road layout, heavy traffic, dense vegetation, as well as complex intersections, Singapore has highly organized multi-lane roads, modern highway layout, organized intersections, and dense vegetation. This results in high variability in scene layout, object distribution, and driving styles. Therefore, it forms a strong foundation for evaluating generalization performance for 3D occupancy prediction.

\begin{figure}[ht]
  \centering
  \includegraphics[width=350pt]{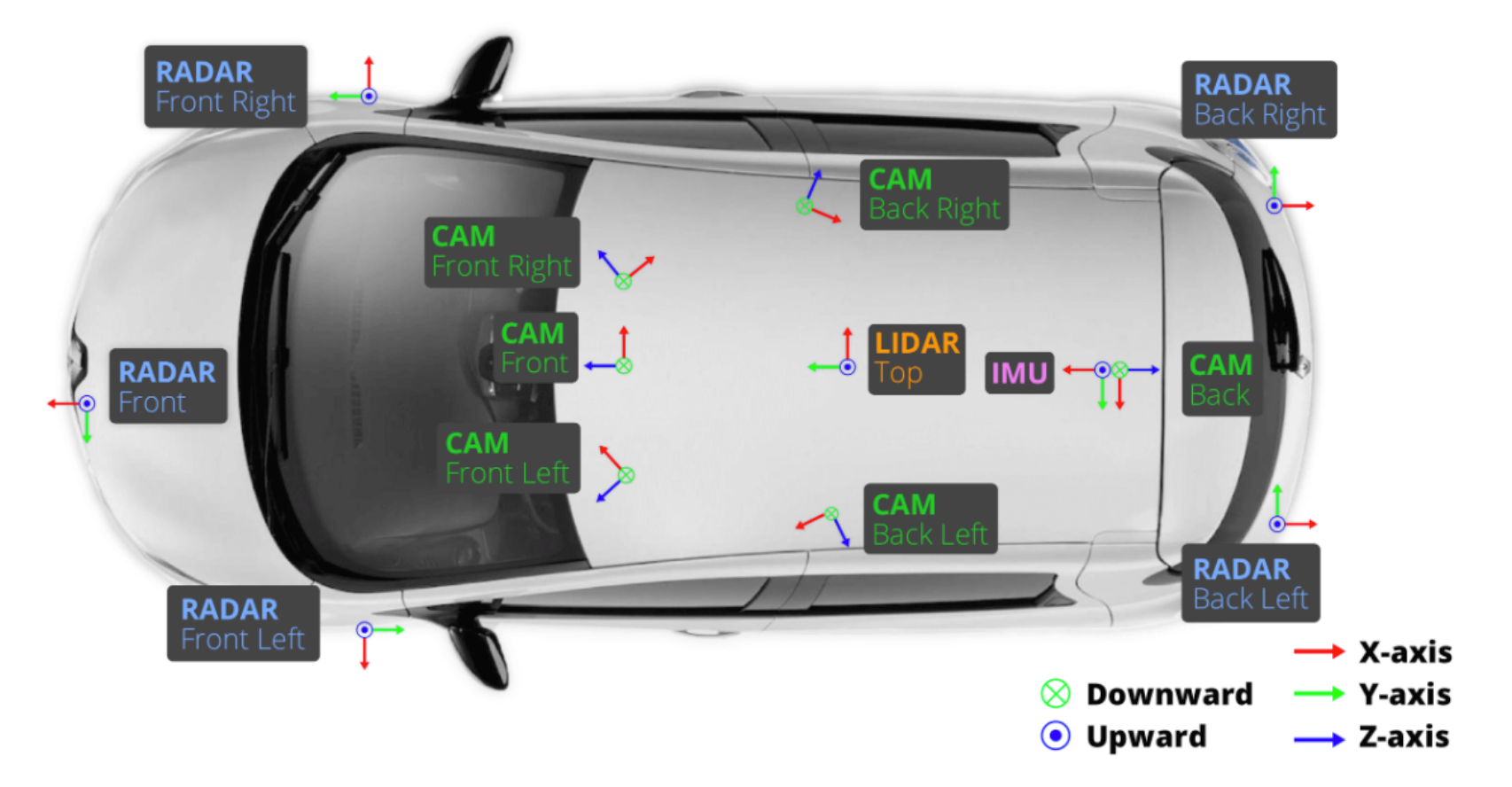}
  \caption[]{Nuscenes sensor setup.}
  \label{fig:nuscenes}
\end{figure}


One of the major advantages offered by the nuScenes dataset is the varied spectrum of weather and lighting conditions it covers. Weather conditions range from clear to overcast, with haze and varied rainfall intensities, coupled with lighting conditions that range from day to sun-set, night, and even nighttime scenarios, and other conditions that include tunnels, wet reflective surfaces, complex roundabouts, construction areas, and vegetation.

The dataset is built around a fully synchronized multi-modal sensor suite, which makes it particularly suitable for camera-LiDAR fusion tasks. The sensor configuration includes:
\begin{itemize}
    \item \textbf{Six wide-field-of-view cameras} providing full $360^\circ$ coverage at a resolution of $1600 \times 900$.
    \item \textbf{One spinning LiDAR} (Velodyne HDL-32E) operating at 20~Hz with dense 3D point measurements.
    \item \textbf{Five RADAR sensors} offering long-range velocity and range information.
    \item \textbf{GPS/IMU} modules enabling precise ego-motion estimation.
    \item Full intrinsic and extrinsic calibration for all sensors.
\end{itemize}


nuScenes dataset is composed of $1,000$ driving scenes lasting $20$ seconds at a rate of about $2$ Hz, thus giving 40,000 keyframes. The dataset distribution is in accordance with the required distribution of $700$ training scenes, $150$ validation scenes, and $150$ scenes to test. nuScenes considers a total of $23$ object categories, but occupancy annotation pipelines map these categories according to their voxel class types.

Since nuScenes does not natively include 3D occupancy labels, we use three widely adopted semantic occupancy annotation pipelines: \textit{SurroundOcc}~\cite{surroundocc}, \textit{Occ3D}~\cite{occ3d}, and \textit{OpenOccupancy}~\cite{openoccupancy2023}. These pipelines transform multi-modal sensor data into semantic voxel grids by applying distinct strategies for LiDAR densification, occlusion handling, and semantic refinement. SurroundOcc provides lightweight and visibility-aware occupancy maps through multi-frame LiDAR aggregation. Occ3D generates refined voxel spaces with 16 semantic categories plus a general-object class, incorporating image-guided completion to improve spatial consistency. OpenOccupancy, the most comprehensive annotation pipeline, produces large-scale voxel grids of size $512 \times 512 \times 40$ with 17 semantic classes using an Augmenting and Purifying Pipeline (AAP). Owing to its scale, completeness, and widespread adoption, OpenOccupancy serves as our primary benchmark for performance evaluation.

Since nuScenes does not provide inherent 3D occupancy annotations, the following top-3 semantic occupancy annotation pipelines have been selected: SurroundOcc~\cite{surroundocc}, Occ3D~\cite{occ3d}, and OpenOccupancy~\cite{openoccupancy2023}. These pipelines transform multi-modal sensor inputs into semantic voxel representation by using different LiDAR densification, occlusion reasoning, and semantic refinement techniques. SurroundOcc provides compact and visibility-aware occupancy representations by aggregating LiDAR points from multiple frames. Occ3D delivers densely voxelized representations of size 16 categories + generalized-object class using image-guided completion to improve the consistency of the representation in the occupied areas. OpenOccupancy, the most detailed occupancy dataset, provides voxel representation of size $512 \times 512 \times 40$ with $17$ semantic categories using the Augmenting and Purifying Pipeline (AAP). Because of the detailed representation characteristics, OpenOccupancy is considered the leading dataset against which performance is compared.

To further analyze robustness under challenging conditions, we construct two additional evaluation subsets from the official validation split. Using the metadata provided by nuScenes, we create:
\begin{itemize}
    \item \textbf{Night subset:} frames recorded during nighttime or low-light conditions.
    \item \textbf{Rain subset:} frames captured under light or moderate rainfall.
\end{itemize}

\subsection{SemanticKITTI Dataset}


The SemanticKITTI~\cite{semantickitti} dataset is introduced as a secondary benchmark to test the generalization skills of the developed 3D occupancy prediction system. The SemanticKITTI dataset is based on the odometry dataset, KITTI, and has been captured in Karlsruhe, a mid-sized city in Germany, which represents a European suburban area with a combination of suburban, residential, and semi-urban settings. The dataset covers a wide spectrum of scenarios, from quiet residential areas, rural areas, industrial areas, curved intersections, to moderately heavily trafficked urban areas, providing a high degree of structural variability to test geometrical consistency and semantic completeness in outdoor scenes.

\begin{figure}[ht]
  \centering
  \includegraphics[width=350pt]{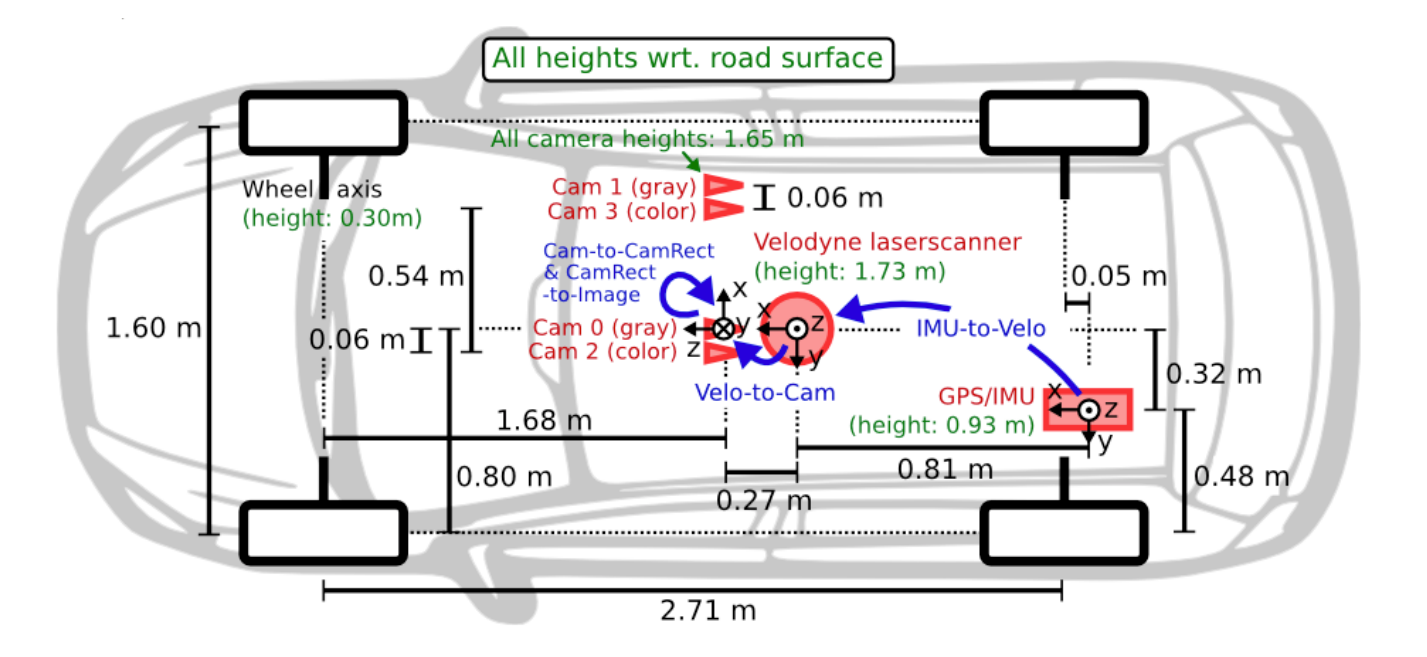}
  \caption[]{SemanticKITTI sensor setup.}
  \label{fig:kitti}
\end{figure}


All sequences in the SemanticKITTI~\cite{semantickitti} dataset are recorded in natural daylight conditions, which include common environmental factors such as cloudy skies, soft shadows caused by vegetation and buildings in the vicinity, and occasional strong sunlight. Even though the dataset does not have weather information or nighttime scenes, it represents a realistic environment of a Central European country, including a combination of vegetation, cars, bicycles, pedestrians, and static scene elements such as buildings, fences, posts, and lines.


SemanticKITTI~\cite{semantickitti} features a single high-resolution Velodyne HDL-64E LiDAR scanner, which provides a rich, dense 360-degree map of points in the scene at around $10$ Hz. Despite the lack of multi-view camera images in the semantic scene completion dataset, the LiDAR measurements provide a rich source of geometric information, and the onboard GPS/IMU system allows high-precision calculation of the vehicle’s ego-motion. It is these factors which make the SemanticKITTI dataset so valuable in the evaluation of voxel-based scene completion algorithms.

The dataset consists of 22 sequences following the official split:
\begin{itemize}
    \item \textbf{Training:} Sequences 00--10 (excluding 08)
    \item \textbf{Validation:} Sequence 08
    \item \textbf{Testing:} Sequences 11--21
\end{itemize}


For occupancy estimation, SemanticKITTI~\cite{semantickitti} offers voxelized semantic annotations on a $256 \times 256 \times 32$ grid with a voxel size of 0.2\,m. Ground truth semantic maps are obtained through LiDAR sweep summation and map refinement techniques, resulting in dense three-dimensional semantic maps. After data preprocessing and class id conversion, a set of 19 semantic classes is used for training and testing, including dynamic objects (vehicles, pedestrians), static objects (buildings, poles, fences), and natural objects (terrain, vegetation).


Though SemanticKITTI~\cite{semantickitti}  lacks an end-to-end camera-LiDAR multimodal setup, its high-res LiDAR geometry, voxel-wise annotations, and well-defined semantic taxonomy make it a useful dataset. It extends the capabilities of the nuScenes dataset, which allows for an examination of generalization capabilities for occupancy prediction methods over LiDAR-centred environments with differing geometric properties.

\section{Implementation Details}


We use a multi-modal approach, utilizing a framework that is based on a Gaussian function, to process both images and LiDAR scans towards 3D occupancy estimation in a combined setting. For testing purposes, experiments are done on three different resolution levels of inputs: 128$\times$256, 256$\times$704, and 900$\times$1600 resolution levels. Otherwise, results will be presented at a resolution of 900$\times$1600 for nuScenes~\cite{nuscenes} and at a resolution of 376$\times$1408 for SemanticKITTI~\cite{kitti360} datasets.


In the case of the camera stream, conventional practices in multi-view 3D perception tasks are considered. In particular, for the nuScenes~\cite{nuscenes} dataset, a ResNet-DCN-101~\cite{resnet} backbone initialized using the FCOS3D~\cite{fcos3d} checkpoint is used, and for SemanticKITTI~\cite{semantickitti}, a ResNet50~\cite{resnet} network pre-trained on ImageNet is used. Furthermore, to achieve multi-scale image feature extraction, a Feature Pyramid Network (FPN)~\cite{fpn} is used in both approaches to extract image features at downsampling ratios of $\{4, 8, 16, 32\}$.


For the LiDAR modality, the analysis covers both sparse and dense settings by adjusting the accumulated sweep count. The sparse setting uses only the LiDAR measurement in the current frame, while the dense setting accumulates up to ten sweeps, depending on the data split. This makes it possible to methodically evaluate the effect of LiDAR density on the accuracy of 3D occupancy maps. The spatial scales follow standard settings: $[-51.2, 51.2]$\,m in the X/Y plane, $[-2.0, 6.0]$\,m in Z for nuScenes~\cite{nuscenes}, while for SemanticKITTI~\cite{semantickitti}, with the forward-facing camera, the LiDAR data captures $[0, 51.2]$\,m in depth and $[-25.6, 25.6]$\,m in the horizontal direction.


Both modalities are integrated into a unified 3D representation through our Gaussian Lifter, which extracts image-based and LiDAR-based features to represent them as continuous Gaussian primitives. Each Gaussian is encapsulated in a learnable mean, covariance, opacity value, and feature embedding. In our main experiments, $12,800$ Gaussians are used for nuScenes~\cite{nuscenes} and $38,400$ Gaussians for SemanticKITTI~\cite{semantickitti}. These Gaussians are further distilled through a four-block Mamba-Gaussian refinement network to enable interaction across image, LiDAR, and position feature modalities.


A Mamba-Transformer decoder incorporating a multi-modal architecture receives the Gaussian layer output. Long range dependencies are captured through selective scanning combined with window attention, enabling the preservation of local geometric structures. Mamba defaults to a feature dimension of 128, of which a 384-dimensional ablation version is also tested.


We train our model using the AdamW optimizer with a weight decay rate of 0.01. The learning rate starts increasing for the first 500 iterations, reaching a maximum of $2\times10^{-4}$, and then applies cosine decay. We perform training for 20 epochs on the nuScenes~\cite{nuscenes} dataset [4] with a batch size of 8 and for 30 epochs on SemanticKITTI~\cite{semantickitti} with a batch size of 4. For evaluation, one can leverage test time augmentation with horizontal flips for increased robustness.


Finally, we offer a thorough evaluation of the effects of image resolution and LiDAR density on the performance of occupancy prediction. Results demonstrate their contributions in enhancing accuracy with the increase in image resolution and LiDAR density, especially in the far areas and boundary of the object.

\section{Evaluation Metrics}

Evaluating a 3D occupancy prediction model on a multiclass voxel grid requires metrics that capture both geometric correctness and semantic reliability across all object categories. Each voxel in the scene is assigned one of $K$ semantic classes (e.g., road, building, vehicle, vegetation). Therefore, an effective evaluation measure must quantify how well the predicted voxel class distribution matches the ground-truth semantic layout in 3D space. In this thesis, we adopt two standard metrics widely used in semantic occupancy and semantic scene completion research: Intersection over Union (IoU) for each semantic class and mean Intersection over Union (mIoU) as the overall performance measure.

\begin{figure}[ht]
  \centering
  \includegraphics[width=250pt]{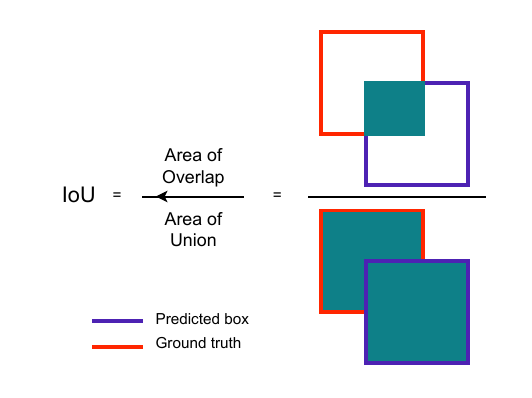}
  \caption[]{Intersection Over Union (IoU).}
  \label{fig:iou}
\end{figure}

\subsection{Intersection over Union (IoU)}

Intersection over Union (IoU) evaluates the quality of predictions on a per-class basis. For each semantic class $c \in \{1, \dots, K\}$, define the predicted voxel set
\begin{equation}
P_c = \{ v \mid \hat{y}(v) = c \},
\end{equation}
and the ground-truth voxel set
\begin{equation}
G_c = \{ v \mid y(v) = c \}.
\end{equation}

The IoU for class $c$ is defined as
\begin{equation}
\text{IoU}_c = 
\frac{|P_c \cap G_c|}
{|P_c \cup G_c|}
=
\frac{|TP_c|}
{|TP_c| + |FP_c| + |FN_c|},
\end{equation}
where $TP_c$, $FP_c$, and $FN_c$ represent the true positives, false positives, and false negatives for class $c$, respectively. An illustration of the intersection and union regions used in IoU computation is shown in Figure~\ref{fig:iou}.

IoU always lies within the interval $[0,1]$. This follows from the fact that the intersection of two sets is always a subset of their union:
\begin{equation}
P_c \cap G_c \subseteq P_c \cup G_c
\quad \Rightarrow \quad
0 \leq \frac{|P_c \cap G_c|}{|P_c \cup G_c|} \leq 1.
\end{equation}

The boundary cases have intuitive interpretations: $\text{IoU}_c = 0$ indicates that no voxel of class $c$ is correctly predicted, while $\text{IoU}_c = 1$ corresponds to perfect agreement between prediction and ground truth for that class. Because semantic voxel grids are highly imbalanced, with large static classes (e.g., road, building) and rare classes (e.g., pedestrians, bicycles), IoU provides a strict and reliable measure of both semantic and geometric consistency.

\subsection{Mean Intersection over Union (mIoU)}

To obtain a single, balanced performance measure across all classes, we compute the mean Intersection over Union (mIoU) as
\begin{equation}
\text{mIoU} = \frac{1}{K} \sum_{c=1}^{K} \text{IoU}_c.
\end{equation}

mIoU assigns equal importance to all semantic classes, regardless of class frequency, making it a fair evaluation metric for multiclass 3D occupancy prediction. This is particularly important because rare classes often occupy only a small percentage of the voxel grid but carry significant meaning for downstream tasks such as navigation and planning. As a result, mIoU is widely used in semantic occupancy datasets such as OpenOccupancy, SemanticKITTI, and SemanticPOSS.

\section{Results}

In this section, we provide a comprehensive quantitative and qualitative evaluation of the GaussianOcc3D framework. We first benchmark our model on the nuScenes~\cite{nuscenes} dataset using both OpenOccupancy~\cite{openoccupancy2023} and Occ3D~\cite{occ3d} annotation standards, followed by a rigorous assessment of generalization on the high-resolution LiDAR scans of SemanticKITTI~\cite{semantickitti}. In order to better analyze the robustness and optimal architectural efficacy of our approach, we conduct a full set of ablation studies, as well as a set of stress tests in challenging environmental settings, such as those in the Rainy and Night conditions. Finally, we analyze the effects of representation density and fusion methods on performance, culminating in a complexity analysis that verifies a good balance of accuracy and complexity. The findings unequivocally show that by using anisotropic gaussian primitives along with adaptive multimodal fusion, a new state of the art in semantic 3D occupancy estimation can be achieved.

\subsection{Results on NuScenes Dataset}


We also analyze the quantitative performance of the proposed GaussianOcc3D approach on the OpenOccupancy~\cite{openoccupancy2023} validation set with the v0.0 annotation scheme for benchmarking the proposed approach against a wide range of the most effective state-of-the-art alternatives for the camera-only, LiDAR-only, and multi-sensor fusion models. It is evident from the results presented in Table~\ref{table_openoccupancy} that the approach establishes a brand-new state-of-the-art for the mean Intersection-over-Union (mIoU) metric of 25.3\%, thus proving more effective for the task of scene reconstruction than any of the previously developed alternatives. Comparing the proposed multi-sensor fusion approach with the alternatives that are restricted to a single sensor, it becomes comprehensible that the applicability of the approach would be significantly better as it outperforms the best performing camera-only network, SparseOcc~\cite{tang2024sparseocc}, of the state of the art with a significantly larger margin of improvement of +11.2\%, and outperforms the best LiDAR-only network, JS3C-Net~\cite{cheng2021s3cnet}, with a relatively larger margin of improvement of +12.8\%.


Within the specific area of multi-modal (Camera \& LiDAR) fusion, the proposed solution retains its strong position in the competitive environment. Compared to the original M-CONet~\cite{openoccupancy2023} and the latest Co-Occ~\cite{coocc2024} solution, the presented technique shows better results by +5.2\% and +3.4\%, correspondingly. Most notably, the technique outperforms the capable OccMamba~\cite{occmamba2024} solution by a large margin and gains the lead position with a +25.2\% increase in total reconstruction accuracy. This progress can be traced on a detailed class-level analysis of IoU values and suggests that the advance occurs not on all but only on those classes that require very high geometric accuracy. The technique shows strong resilience on the reconstruction tasks related to small, thin, and irregular objects. These categories are specifically problematic for the voxel-based approaches that suffer from quantization issues. In particular, the solution shows the absolute best results on the Barriers (29.7\%), Bicycles (19.8\%), Motorcycles (25.5\%), and Traffic Cones (19.9\%) categories. This improvement can be explained by the anisotropic properties of the used Gaussian primitives and the tendency of the GaussianOcc3D solution to better fit the thin shape of the bicycle and the shape of the traffic cone compared to the competition.


The model showcases its strength in dealing with extensive static context and dynamic entities successfully. To be precise, the GaussianOcc3D achieves the best accuracy of 28.9\% on Cars and 24.7\% on Trucks, signifying the success of the feature aggregation component in incorporating the detailed textures obtained from cameras and the geometric info of LiDAR in accurately defining the boundary of vehicles. On the context front, the solution sets a new state of the art on the ground plane estimation task with the best score of 37.5\% on Driveable Surface and 25.9\% on Terrain. A fine segmentation of the driveable surface is extremely important in the context of autonomous driving, and the continuous representation of Gaussian splats is successful in representing the plane geometry without the typical staircase effect of voxel representation.

\definecolor{barriercolor}{RGB}{255, 0, 0}
\definecolor{bicyclecolor}{RGB}{0, 255, 0}
\definecolor{buscolor}{RGB}{255, 255, 0}
\definecolor{carcolor}{RGB}{0, 0, 255}
\definecolor{construction_vehiclecolor}{RGB}{255, 165, 0}
\definecolor{motorcyclecolor}{RGB}{128, 0, 128}
\definecolor{pedestriancolor}{RGB}{255, 192, 203}
\definecolor{traffic_conecolor}{RGB}{255, 69, 0}
\definecolor{trailercolor}{RGB}{192, 192, 192}
\definecolor{truckcolor}{RGB}{139 ,69 ,19}
\definecolor{driveable_surfacecolor}{RGB}{135 ,206 ,235}
\definecolor{other_flatcolor}{RGB}{160 ,82 ,45}
\definecolor{sidewalkcolor}{RGB}{211 ,211 ,211}
\definecolor{terraincolor}{RGB}{139 ,105 ,20}
\definecolor{manmadecolor}{RGB}{112 ,128 ,144}
\definecolor{vegetationcolor}{RGB}{34 ,139 ,34}

\newcommand{\makecolorcell}[2]{%
    \makecell[b]{\begin{turn}{90}\colorbox{#1}{\hspace{3pt}\rule{0pt}{3pt}} #2\end{turn}}}

\begin{table}[htbp]
\centering
\resizebox{\linewidth}{!}{
\setlength{\tabcolsep}{1mm}
\begin{threeparttable}
\begin{tabular}{l|c|c|cccccccccccccccc}
\toprule
\makecell[b]{Method} & \makecell[b]{Modality} & 
\makecell[b]{mIoU} & 
\makecolorcell{barriercolor}{barrier} & 
\makecolorcell{bicyclecolor}{bicycle} & 
\makecolorcell{buscolor}{bus} & 
\makecolorcell{carcolor}{car} & 
\makecolorcell{construction_vehiclecolor}{const. veh.} & 
\makecolorcell{motorcyclecolor}{motorcycle} & 
\makecolorcell{pedestriancolor}{pedestrian} & 
\makecolorcell{traffic_conecolor}{traffic cone} & 
\makecolorcell{trailercolor}{trailer} & 
\makecolorcell{truckcolor}{truck} & 
\makecolorcell{driveable_surfacecolor}{drive surf.} & 
\makecolorcell{other_flatcolor}{other\_flat} & 
\makecolorcell{sidewalkcolor}{sidewalk} & 
\makecolorcell{terraincolor}{terrain} & 
\makecolorcell{manmadecolor}{manmade} & 
\makecolorcell{vegetationcolor}{vegetation} \\

\midrule

MonoScene~\cite{cao2022monoscene} & C & 6.9 & 7.1 & 3.9 & 9.3 & 7.2 & 5.6 & 3.0 & 5.9 & 4.4 & 4.9 & 4.2 & 14.9 & 6.3 & 7.9 & 7.4 & 10.0 & 7.6 \\
TPVFormer~\cite{yuan2023tpvformer} & C & 7.8 & 9.3 & 4.1 & 11.3 & 10.1 & 5.2 & 4.3 & 5.9 & 5.3 & 6.8 & 6.5 & 13.6 & 9.0 & 8.3 & 8.0 & 9.2 & 8.2 \\
SparseOcc~\cite{tang2024sparseocc} & C & 14.1 & 16.1 & 9.3 & 15.1 & 18.6 & 7.3 & 9.4 & 11.2 & 9.4 & 7.2 & 13.0 & 31.8 & 21.7 & 20.7 & 18.8 & 6.1 & 10.6 \\
\midrule
3DSketch~\cite{mitani20003d} & C\&D & 10.7 & 12.0 & 5.1 & 10.7 & 12.4 & 6.5 & 4.0 & 5.0 & 6.3 & 8.0 & 7.2 & 21.8 & 14.8 & 13.0 & 11.8 & 12.0 & 21.2 \\
\midrule
LMSCNet~\cite{roldao2020lmscnet} & L & 11.5 & 12.4 & 4.2 & 12.8 & 12.1 & 6.2 & 4.7 & 6.2 & 6.3 & 8.8 & 7.2 & 24.2 & 12.3 & 16.6 & 14.1 & 13.9 & 22.2 \\
JS3C-Net~\cite{cheng2021s3cnet} & L & 12.5 & 14.2 & 3.4 & 13.6 & 12.0 & 7.2 & 4.3 & 7.3 & 6.8 & 9.2 & 9.1 & 27.9 & 15.3 & 14.9 & 16.2 & 14.0 & 24.9 \\
\midrule
M-CONet~\cite{openoccupancy2023} & C\&L & 20.1 & 23.3 & 13.3 & 21.2 & 24.3 & 15.3 & 15.9 & 18.0 & 13.3 & 15.3 & 20.7 & 33.2 & 21.0 & 22.5 & 21.5 & 19.6 & 23.2 \\
Co-Occ~\cite{coocc2024} & C\&L & 21.9 & 26.5 & 16.8 & 22.3 & 27.0 & 10.1 & 20.9 & 20.7 & 14.5 & 16.4 & 21.6 & 36.9 & 23.5 & 5.5 & 23.7 & 20.5 & 23.5 \\
OccMamba~\cite{occmamba2024} & C\&L & \underline{25.2} & \underline{29.1} & \underline{19.1} & \textbf{25.5} & \underline{28.5} & \textbf{18.1} & \underline{24.7} & \textbf{23.4} & \underline{19.8} & \textbf{19.3} & \underline{24.5} & \underline{37.0} & \textbf{25.4} & \textbf{25.4} & \underline{25.4} & \textbf{28.1} & \textbf{29.9} \\

\rowcolor{lightgray!45} \textbf{GaussianOcc3D (ours)} & C\&L & \textbf{25.3} & \textbf{29.7} & \textbf{19.8} & 25.1 & \textbf{28.9} & \underline{17.8} & \textbf{25.5} & \underline{23.0} & \textbf{19.9} & \underline{19.1} & \textbf{24.7} & \textbf{37.5} & \underline{25.1} & \underline{25.3} & \textbf{25.9} & \underline{27.8} & \underline{29.7} \\

\bottomrule
\end{tabular}
\end{threeparttable}
}
\caption[]{3D semantic occupancy prediction results on OpenOccupancy validation set.}
\label{table_openoccupancy}
\vspace{-3mm}
\end{table}


To better analyze the generalization and robustness of the new strategy introduced in the paper, we conduct comprehensive experiments on the Occ3D-nuScenes validation set, a large-scale benchmark that exhibits strong geometric complexity and refined semantic annotation. The comparative results against a variety of strong baselines ranging from vision Transformers to the latest multi-modal fusion architectures are summarized in Table \ref{tab:occ3d}. Our new GaussianOcc3D system produces state-of-the-art mIoU scores of 49.4\%, a significant boost to the current state-of-the-art. This new mIoU score not only sets a new record but also indicates a steady gain of +0.7\% and +3.0\% against the current state-of-the-art multi-modal OccFusion~\cite{occfusionming2024} and GaussianFormer3D~\cite{gaussianformer3d2025} systems. Most notably, compared to the latest camera-based strong baseline COTR~\cite{ma2024cotr} (mIoU=44.5\%), a nearly +5\% boost to the mIoU can be obtained with the new multi-modal system introduced in the paper. This clearly asserts the valuable contribution from depth-related terms toward alleviating the inherent geometric ambiguities in image-based reconstruction tasks.


Class-wise analysis of results highlights the unique architectural strengths of our approach enabled by the representation using Gaussians. The algorithm performs exceptionally well in reconstructing unstructured scenes. In the Vegetation class, which involves modeling complex, porous surfaces like tree branches and shrubs, our algorithm, GaussianOcc3D, records a score of 69.5\%, outscoring OccFusion by a substantial margin of 8.2 percentage points, reporting a score of 61.3\%. Similarly, in the category of Manmade scenes, our algorithm records the best score of 69.7\%, significantly bettering voxel-based competitors. These results provide credence to our assumption that anisotropic primitives described by Gaussians are naturally more adept at modeling continuous irregular surfaces compared to rigid voxel grids, which tend to introduce artifacts of discretization when modeling irregular surfaces.


The proposed method demonstrates excellent performance on dynamic and safety-constrained object categories. It attains the highest IoU values of 59.7\% on the Car category, 34.5\% on the Bus category, and 43.7\% on the Motorcycle category. Its outstanding performance on the Motorcycle category, which has thin structure components, demonstrates the accuracy of the depth-feature lifting process, as the method successfully maintains the detailed structure information of the motorcycle, which is often lost using traditional aggregation operations. In addition, the proposed method attains the best performance on small object detection benchmarks, where the Traffic Cone category reaches 48.6\%, showing the ability of the multi-depth attention mechanism to distinguish the small obstacle from the background in sparse LiDAR points. Despite the merits of the COTR~\cite{ma2024cotr} algorithm on certain plane categories (e.g., the Driveable Surface category), the GaussianOcc3D algorithm presents the most well-rounded performance over the whole range of semantic categories, thus confirming the benefit of the bifurcated multi-modal fusion solution.

\definecolor{nbarrier}{RGB}{255, 120, 50}
\definecolor{nbicycle}{RGB}{255, 192, 203}
\definecolor{nbus}{RGB}{255, 255, 0}
\definecolor{ncar}{RGB}{0, 150, 245}
\definecolor{nconstruct}{RGB}{0, 255, 255}
\definecolor{nmotor}{RGB}{200, 180, 0}
\definecolor{npedestrian}{RGB}{255, 0, 0}
\definecolor{ntraffic}{RGB}{255, 240, 150}
\definecolor{ntrailer}{RGB}{135, 60, 0}
\definecolor{ntruck}{RGB}{160, 32, 240}
\definecolor{ndriveable}{RGB}{255, 0, 255}
\definecolor{nother}{RGB}{139, 137, 137}
\definecolor{nsidewalk}{RGB}{75, 0, 75}
\definecolor{nterrain}{RGB}{150, 240, 80}
\definecolor{nmanmade}{RGB}{213, 213, 213}
\definecolor{nvegetation}{RGB}{0, 175, 0}\definecolor{tan}{rgb}{0.82, 0.71, 0.55}

\begin{table}[htbp]
  \centering
  \resizebox{\linewidth}{!}{
    \begin{tabular}{l|c|c|ccccccccccccccccc}
      \toprule
      Method & Modality  & mIoU & 
      \rotatebox{90}{others} &
      \rotatebox{90}{barrier} & 
      \rotatebox{90}{bicycle} & 
      \rotatebox{90}{bus} & 
      \rotatebox{90}{car} & 
      \rotatebox{90}{const. veh.} & 
      \rotatebox{90}{motorcycle} & 
      \rotatebox{90}{pedestrian} & 
      \rotatebox{90}{traffic cone} & 
      \rotatebox{90}{trailer} & 
      \rotatebox{90}{truck} & 
      \rotatebox{90}{drive. surf.} & 
      \rotatebox{90}{other flat} & 
      \rotatebox{90}{sidewalk} & 
      \rotatebox{90}{terrain} & 
      \rotatebox{90}{manmade} & 
      \rotatebox{90}{vegetation} \\
      &  & & 
      \tikz \draw[fill=black,draw=black] (0,0) rectangle (0.2,0.2); &       
      \tikz \draw[fill=nbarrier,draw=nbarrier] (0,0) rectangle (0.2,0.2); & 
      \tikz \draw[fill=nbicycle,draw=nbicycle] (0,0) rectangle (0.2,0.2); & 
      \tikz \draw[fill=nbus,draw=nbus] (0,0) rectangle (0.2,0.2); & 
      \tikz \draw[fill=ncar,draw=ncar] (0,0) rectangle (0.2,0.2); & 
      \tikz \draw[fill=nconstruct,draw=nconstruct] (0,0) rectangle (0.2,0.2); & 
      \tikz \draw[fill=nmotor,draw=nmotor] (0,0) rectangle (0.2,0.2); & 
      \tikz \draw[fill=npedestrian,draw=npedestrian] (0,0) rectangle (0.2,0.2); & 
      \tikz \draw[fill=ntraffic,draw=ntraffic] (0,0) rectangle (0.2,0.2); & 
      \tikz \draw[fill=ntrailer,draw=ntrailer] (0,0) rectangle (0.2,0.2); & 
      \tikz \draw[fill=ntruck,draw=ntruck] (0,0) rectangle (0.2,0.2); & 
      \tikz \draw[fill=ndriveable,draw=ndriveable] (0,0) rectangle (0.2,0.2); & 
      \tikz \draw[fill=nother,draw=nother] (0,0) rectangle (0.2,0.2); & 
      \tikz \draw[fill=nsidewalk,draw=nsidewalk] (0,0) rectangle (0.2,0.2); & 
      \tikz \draw[fill=nterrain,draw=nterrain] (0,0) rectangle (0.2,0.2); & 
      \tikz \draw[fill=nmanmade,draw=nmanmade] (0,0) rectangle (0.2,0.2); & 
      \tikz \draw[fill=nvegetation,draw=nvegetation] (0,0) rectangle (0.2,0.2); \\
      \midrule
      MonoScene~\cite{cao2022monoscene} & C & 6.1& 1.8& 7.2& 4.3& 4.9& 9.4& 5.7& 4.0& 3.0& 5.9& 4.5& 7.2& 14.9& 6.3& 7.9& 7.4& 1.0& 7.7 \\
      BEVDet~\cite{huang2021bevdet} & C & 11.7& 2.1& 15.3& 0.0& 4.2& 13.0& 1.4& 0.0& 0.4& 0.13& 6.6& 6.7& 52.7& 19.0& 26.5& 21.8& 14.5& 15.3\\
      BEVFormer~\cite{li2024bevformer} & C & 23.7& 5.0& 38.8& 10.0& 34.4& 41.1& 13.2& 16.5& 18.2& 17.8& 18.7& 27.7& 49.0& 27.7& 29.1& 25.4& 15.4& 14.5\\
      TPVFormer~\cite{yuan2023tpvformer} & C & 28.3& 6.7& 39.2& 14.2& 41.5& 47.0& 19.2& 22.6& 17.9& 14.5& 30.2& 35.5& 56.2& 33.7& 35.7& 31.6& 20.0& 16.1\\
      RenderOcc~\cite{pan2024renderocc} & C & 26.1& 4.8& 31.7& 10.7& 27.7& 26.5& 13.9& 18.2& 17.7& 17.8& 21.2& 23.3& 63.2& 36.4& 46.2& 44.3& 19.6& 20.7\\
      GaussianFormer~\cite{huang2024gaussianformer} & C & 35.5& 8.8 & 40.9 & 23.3 & 42.9 & 49.7 & 19.2 & 24.8 & 24.4 & 22.5 & 29.4 & 35.3 & 79.0 & 36.9 & 46.6 & 48.2 & 38.8 & 33.1 \\
      BEVDet4D~\cite{huang2022bevdet4d} & C & 39.3 & 9.3& 47.1& 19.2& 41.5& 52.2& 27.2& 21.2& 23.3& 21.6& 35.8& 38.9& 82.5& 40.4& 53.8& 57.7& 49.9& 45.8 \\
      COTR~\cite{ma2024cotr} & C & 44.5& \underline{13.3}& \underline{52.1}& 32.0& 46.0& 55.6& \underline{32.6}& 32.8& 30.4& \underline{34.1}& 37.7& 41.8& \textbf{84.5}& \underline{46.2}& \textbf{57.6}& \underline{60.7}& 52.0& 46.3\\
      PanoOcc~\cite{openoccupancy2023} & C & 42.1& 11.7& 50.5& 29.6& 49.4& 55.5& 23.3& 33.3& 30.6& 31.0& 34.4& 42.6& \underline{83.3}& 44.2& 54.4& 56.0& 45.9& 40.4\\
      FB-Occ~\cite{li2023fbocc} & C & 42.1 & \textbf{14.3}& 49.7& 30.0& 46.6& 51.5& 29.3& 29.1& 29.4& 30.5& 35.0& 39.4& 83.1& \textbf{47.2}& 55.6& 59.9& 44.9& 39.6\\
      \midrule
      OccFusion~\cite{occfusionming2024} & C\&L & \underline{48.7} & 12.4 & 51.8 & \underline{33.0}& \underline{54.6}& 57.7 & \textbf{34.0} & \underline{43.0} & \underline{48.4} & \textbf{35.5}& \textbf{41.2}& 48.6& 83.0& 44.7& \underline{57.1}& 60.0& 62.5& 61.3 \\
      GaussianFormer3D~\cite{gaussianformer3d2025} & C\&L & 46.4 & 9.8 & 50.0 & 31.3 & 54.0 & \underline{59.4} & 28.1 & 36.2 & 46.2 & 26.7 & 40.2 &  \textbf{49.7} & 79.1 & 37.3 & 49.0 & 55.0 & \underline{69.1} & \underline{67.6}\\
      \rowcolor{lightgray!45}  \textbf{GaussianOcc3D (ours)} & C\&L & \textbf{49.4} & 12.5 & \textbf{54.2} & \textbf{34.5} & \textbf{56.4} & \textbf{59.7} & 32.4 & \textbf{43.7} & \textbf{48.6} & 32.7 & \underline{40.8} &  \underline{49.5} & 81.1 & 40.3 & 54.1 & \textbf{60.9} & \textbf{69.7} & \textbf{69.5}\\
      
      \bottomrule
    \end{tabular}
  }
  \caption[]{3D semantic occupancy prediction results on Occ3D validation set.}
  \label{tab:occ3d}
\end{table}

\subsection{Results on SemanticKITTI Dataset}


To better analyze the robustness of GaussianOcc3D with diverse sensor configurations and densities in the scene, we conduct experiments on the SemanticKITTI test dataset as well. Unlike OpenOccupancy and Occ3D datasets constructed from nuScenes, the SemanticKITTI dataset is captured by a single high-density LiDAR sensor (Velodyne HDL-64E) resulting in a different point cloud pattern with greater geometric detail but much smaller vertical fields of view. As shown in Table~\ref{table_semantickitti_full}, state-of-the-art mean Intersection-over-Union (mIoU) performance of 25.2\% is achieved by our framework, outperforming all single-modal methods as well as existing multi-modal methods. A quantitative analysis shows that the severe shortcomings of vision-based approaches in this task are clearly evident; existing approaches like MonoScene~\cite{cao2022monoscene} and OccFormer~\cite{zhang2023occformer} record mIoUs only of 11.1\% and 12.3\%, respectively, and lack the capability to recover adequate geometric structure from sparse views of the cameras without access to explicit depth data. While LiDAR-based solutions like JS3C-Net~\cite{cheng2021s3cnet} perform better (mIoU = 23.8\%), they affirm the crucial roles of geometric priors in this dataset. But interestingly, our fusion method outperforms the strongest LiDAR-centric competitor (SSC-RS~\cite{mei2023ssc}) by a whole percentage index (+1.0\%), thus proving that the addition of high-density semantic texture captured from cameras acts as a categorical complement to the high geometric resolution of LiDAR.


On the challenging front of the state-of-the-art multi-modal fusion, GaussianOcc3D eclipses the former state-of-the-art OccMamba~\cite{occmamba2024} and thus sets a new state of the art. The granular result per class points to the fact that the architectural benefits lie in the unstructured and slender geometry regime. Our approach sets a new state of the art in the Vegetation (45.6\%) category, Fence (32.8\%), and Pole (27.1\%) categories simultaneously. All these categories represent the most geometrically complex scene components: the vegetation refers to the representation of the irregular, porous volumes, and the representation of the high-aspect-ratio, thin structures related to the fences and poles. The respective better performance is directly related to the fact that we have the anisotropic representation ability of the Gaussian primitives in our case. By contrast, the method of OccMamba~\cite{occmamba2024} or M-CONet~\cite{openoccupancy2023} has to resort to using the stair-case representation of the slanted fences or the tree branches, which would be impossible by using the cube representation in any reasonable way.


In addition, our model shows strong sensitivity to vulnerable road users, as well as smaller objects. To be more precise, our method achieves the second-best accuracy on Person with 5.1\% accuracy, as well as on Bicyclist with 0.6\% accuracy, while being competitive on Traffic Sign with 23.1\% accuracy. Although the baseline SSC-RS~\cite{mei2023ssc} outperforms our model by a small margin on large planar classes, such as Road with 3.1\% accuracy, as well as on Parking with 1.1\% accuracy, our model, with its representation bridging the semantic gap between different types of sensors, shows more balanced accuracy over the semantic spectrum. Our method shows strong robustness on a large spectrum of scenarios, varying from the organized city scenarios in nuScenes~\cite{nuscenes} datasets, to the more rural environments in SemanticKITTI~\cite{semantickitti} datasets.

\definecolor{roadcolor}{RGB}{255, 40, 200}
\definecolor{sidewalkcolor}{RGB}{240, 20, 255}
\definecolor{parkingcolor}{RGB}{255, 150, 250}
\definecolor{other-groundcolor}{RGB}{175, 0, 75}
\definecolor{buildingcolor}{RGB}{255, 200, 0}
\definecolor{carcolor}{RGB}{245, 150, 100}
\definecolor{truckcolor}{RGB}{180, 30, 80}
\definecolor{bicyclecolor}{RGB}{250, 80, 100}
\definecolor{motorcyclecolor}{RGB}{150, 60, 30}
\definecolor{other-vehiclecolor}{RGB}{255, 0, 0}
\definecolor{vegetationcolor}{RGB}{0, 175, 0}
\definecolor{trunkcolor}{RGB}{135, 60, 0}
\definecolor{terraincolor}{RGB}{150, 240, 80}
\definecolor{personcolor}{RGB}{30, 30, 255}
\definecolor{bicyclistcolor}{RGB}{200, 40, 255}
\definecolor{motorcyclistcolor}{RGB}{90, 30, 150}
\definecolor{fencecolor}{RGB}{255, 120, 50}
\definecolor{polecolor}{RGB}{255, 192, 203}
\definecolor{traffic-signcolor}{RGB}{255, 255, 0}

\begin{table}[htbp]
\centering
\fontsize{8pt}{10pt}\selectfont
\setlength{\tabcolsep}{0.4mm}
\begin{threeparttable}
\begin{tabular}{l|c|c|ccccccccccccccccccc} 
\toprule

Method & \makecell[b]{Modality} & \makecell[b]{mIoU} & \makecell[b]{\begin{turn}{90}\colorbox{roadcolor}{\hspace{3pt}\rule{0pt}{3pt}} road\end{turn}} & \makecell[b]{\begin{turn}{90}\colorbox{sidewalkcolor}{\hspace{3pt}\rule{0pt}{3pt}} sidewalk\end{turn}} & \makecell[b]{\begin{turn}{90}\colorbox{parkingcolor}{\hspace{3pt}\rule{0pt}{3pt}} parking\end{turn}} & \makecell[b]{\begin{turn}{90}\colorbox{other-groundcolor}{\hspace{3pt}\rule{0pt}{3pt}} other ground\end{turn}} & \makecell[b]{\begin{turn}{90}\colorbox{buildingcolor}{\hspace{3pt}\rule{0pt}{3pt}} building\end{turn}} & \makecell[b]{\begin{turn}{90}\colorbox{carcolor}{\hspace{3pt}\rule{0pt}{3pt}} car\end{turn}} & \makecell[b]{\begin{turn}{90}\colorbox{truckcolor}{\hspace{3pt}\rule{0pt}{3pt}} truck\end{turn}} & \makecell[b]{\begin{turn}{90}\colorbox{bicyclecolor}{\hspace{3pt}\rule{0pt}{3pt}} bicycle\end{turn}} & \makecell[b]{\begin{turn}{90}\colorbox{motorcyclecolor}{\hspace{3pt}\rule{0pt}{3pt}} motorcycle\end{turn}} & \makecell[b]{\begin{turn}{90}\colorbox{other-vehiclecolor}{\hspace{3pt}\rule{0pt}{3pt}} other vehicle\end{turn}} & \makecell[b]{\begin{turn}{90}\colorbox{vegetationcolor}{\hspace{3pt}\rule{0pt}{3pt}} vegetation\end{turn}} & \makecell[b]{\begin{turn}{90}\colorbox{trunkcolor}{\hspace{3pt}\rule{0pt}{3pt}} trunk\end{turn}} & \makecell[b]{\begin{turn}{90}\colorbox{terraincolor}{\hspace{3pt}\rule{0pt}{3pt}} terrain\end{turn}} & \makecell[b]{\begin{turn}{90}\colorbox{personcolor}{\hspace{3pt}\rule{0pt}{3pt}} person\end{turn}} & \makecell[b]{\begin{turn}{90}\colorbox{bicyclistcolor}{\hspace{3pt}\rule{0pt}{3pt}} bicyclist\end{turn}} & \makecell[b]{\begin{turn}{90}\colorbox{motorcyclistcolor}{\hspace{3pt}\rule{0pt}{3pt}} motorcyclist\end{turn}} & \makecell[b]{\begin{turn}{90}\colorbox{fencecolor}{\hspace{3pt}\rule{0pt}{3pt}} fence\end{turn}} & \makecell[b]{\begin{turn}{90}\colorbox{polecolor}{\hspace{3pt}\rule{0pt}{3pt}} pole\end{turn}} & \makecell[b]{\begin{turn}{90}\colorbox{traffic-signcolor}{\hspace{3pt}\rule{0pt}{3pt}} traffic sign\end{turn}} \\

\midrule

MonoScene~\cite{cao2022monoscene} & C & 11.1 & 54.7 & 27.1 & 24.8 & 5.7 & 14.4 & 18.8 & 3.3 & 0.5 & 0.7 & 4.4 & 14.9 & 2.4 & 19.5 & 1.0 & 1.4 & 0.4 & 11.1 & 3.3 & 2.1 \\
SurroundOcc~\cite{surroundocc} & C & 11.9 & 56.9 & 28.3 & 30.2 & 6.8 & 15.2 & 20.6 & 1.4 & 1.6 & 1.2 & 4.4 & 14.9 & 3.4 & 19.3 & 1.4 & 2.0 & 0.1 & 11.3 & 3.9 & 2.4 \\
OccFormer~\cite{zhang2023occformer} & C & 12.3 & 55.9 & 30.3 & 31.5 & 6.5 & 15.7 & 21.6 & 1.2 & 1.5 & 1.7 & 3.2 & 16.8 & 3.9 & 21.3 & 2.2 & 1.1 & 0.2 & 11.9 & 3.8 & 3.7 \\
RenderOcc~\cite{pan2024renderocc} & C & 12.8 & 57.2 & 28.4 & 16.1 & 0.9 & 18.2 & 24.9 & 6.0 & 0.4 & 0.3 & 3.7 & 26.2 & 4.9 & 3.6 & 1.9 & 3.1 & 0.0 & 9.1 & 6.2 & 3.4 \\
\midrule
LMSCNet~\cite{roldao2020lmscnet} & L & 17.0 & 64.0 & 33.1 & 24.9 & 3.2 & 38.7 & 29.5 & 2.5 & 0.0  & 0.0 & 0.1 & 40.5 & 19.0 & 30.8 & 0.0 & 0.0 & 0.0 & 20.5 & 15.7 & 0.5 \\
JS3C-Net~\cite{cheng2021s3cnet} & L & 23.8 & 64.0 & 39.0 & 34.2 & \underline{14.7} & 39.4 & 33.2 & 7.2 & \textbf{14.0} & \textbf{8.1} & \textbf{12.2} & 43.5 & 19.3 & 39.8 & \textbf{7.9} & \textbf{5.2} & 0.0 & 30.1 & 17.9 & 15.1 \\
SSC-RS~\cite{mei2023ssc} & L & 24.2 & \textbf{73.1} & \textbf{44.4} & \underline{38.6} & \textbf{17.4} & \textbf{44.6} & \underline{36.4} & 5.3 & \underline{10.1} & 5.1 & \underline{11.2} & 44.1 & 26.0 & \textbf{41.9} & 4.7 & 2.4 & \textbf{0.9} & 30.8 & 15.0 & 7.2 \\
\midrule
Co-Occ~\cite{coocc2024} & C\&L & 24.4 & \underline{72.0} & \underline{43.5} & \textbf{42.5} & 10.2 & 35.1 & \textbf{40.0} & 6.4 & 4.4 & 3.3 & 8.8 & 41.2 & \textbf{30.8} & 40.8 & 1.6 & \underline{3.3} & 0.4 & \underline{32.7} & 26.6 & 20.7 \\
M-CONet~\cite{openoccupancy2023} & C\&L & 20.4 & 60.6 & 36.1 & 29.0 & 13.0 & 38.4 & 33.8 & 4.7 & 3.0 & 2.2 & 5.9 & 41.5 & 20.5 & 35.1 & 0.8 & 2.3 & \underline{0.6} & 26.0 & 18.7 & 15.7 \\

OccMamba~\cite{occmamba2024} & C\&L & \underline{24.6} & 68.7 & 41.0 & 35.9 & 9.1 & 40.8 & 34.8 & \underline{8.8} & 8.8 & 6.5 & 8.9 & \underline{44.9} & \underline{28.7} & 40.6 & 4.2 & 2.6 & \underline{0.6} & 32.0 & \underline{27.0} & \textbf{23.3} \\

\rowcolor{lightgray!45}  \textbf{GaussianOcc3D (ours)} & C\&L & \textbf{25.2} & 69.3& 41.8& 37.8& 10.7& \underline{42.9}& 35.6& \underline{8.7}& \underline{8.9}& \underline{6.9}& 9.2& \textbf{45.6}& 28.2& \underline{41.1}& \underline{5.1}& 2.8& \underline{0.6} & \textbf{32.8} & \textbf{27.1} & \underline{23.1} \\

\bottomrule
\end{tabular}
\end{threeparttable}
\caption[]{3D semantic occupancy prediction results on SemanticKITTI test set.}
\label{table_semantickitti_full}
\end{table}

\subsection{Ablation Study}


For testing the effectiveness of individual constituent elements proposed in the framework, a thorough ablation study has been conducted on the Occ3D-nuScenes validation set. Results from quantitative experiments, as presented in Table~\ref{table:abb}, show a monotonic improvement in mean Intersection-over-Union (mIoU) values while sequentially adding each module to the network, which implies focusing on separate bottlenecks in the perception pipeline for our proposed solutions.

\textbf{Impact of Multi-Modal Integration:} The baseline is set using the gaussianformer with the camera modality alone, achieving an mIoU of 35.5\%. The relatively poor performance emphasizes the ill-posed nature of the task of inferring three-dimensional geometry from two-dimensional projections, with the inherent depth uncertainty resulting in large occupancy prediction errors. Adding the LiDAR modality with a conventional concatenation fusion approach yields the biggest individual boost, outperforming the baseline by a margin of 8.9 percentage points and achieving an mIoU of 44.4\%. These experiments empirically confirm the key premise laid down: geometric priors are crucial for reliable occupancy estimation, providing the needed depth information not captured solely with vision modalities.

\textbf{Effectiveness of Adaptive Fusion (ADA-Fusion):} Using the proposed Adaptive Camera-LiDAR Fusion module in place of the naive concatenation baseline further improves the model’s performance by 1.8 percentage points, reaching 46.2\% mIoU. This improvement is due to the effectiveness brought by its Consistency-Aware Reweighting and Dual-Stream Cross-Attention mechanisms. Compared to naive fusion, which directly combines features uniformly, the proposed module actively alleviates conflicts existing in both sensors, such as reflection and calibration noise. This empirical verification not only affirms the importance of modality choice, equally important to the choice of modalities, but also manifests its effectiveness in suppressing ghosting artifacts before they affect the latent features.

\textbf{Efficacy of LiDAR Depth Feature Aggregation (LDFA):} Adding in the LDFA module provides a significant improvement of +2.1 percentage points, boosting mIoU to 48.3\%. This specific result has key significance in light of its relevance to dealing with the sparsity pattern, a LiDAR data inherent property. Indeed, LiDAR data is traditionally handled in a similar fashion to regular pixels from cameras, leading to a tendency for features to spread through sparse voxels in a scattered pattern. Nonetheless, through Depth-Wise Deformable Sampling and Stochastic Depth Partitioning, LDFA allows for better separation based on LiDAR points, emphasizing actual geometries from empty spaces, which gives clear indications that there is optimization in boosting actual geometries on top of Gaussian primitives without being clouded by volume noise.

\textbf{Role of Entropy-Based Smoothing:} Adding entropy-based feature smoothing brings a steady boost, raising the mIoU to 48.7\%. While the quantitative boost of +0.4\% appears somewhat small compared to gains from the structural modules, the role of this component is mostly one of regularization. Through the estimation of cross-entropy in the opposite directions, the identification and compensation of domain-specific information differences, such as shadow information apparent only in specific camera views, helps to suppress negative transfer in the process. This specific gain helps to confirm that in dynamically equalizing feature distributions before combining them, the learned representation can be made more robust.

\textbf{Contribution of the Gauss-Mamba Head:} At last, by replacing the traditional decoding head with the proposed Gauss-Mamba Head, the model attains its peak performance at 49.4\% mIoU (+0.7\%). This very small boost on top of the strong baseline confirms that the proposed Tri-Perspective View (TPV)~\cite{yuan2023tpvformer} decomposition and the Selective State Space Model indeed work well together. While traditional attention-based methods struggle with modeling the global context because of quadratic complexity, the Mamba head efficiently handles the long-distance dependencies (such as road connectivity) with linear complexity. It can be seen that improving the geometric properties of the Gaussian primitives through a context-conscious and globally informed process is better than doing so individually.

\begin{table}[htbp]

\centering
\scalebox{0.8}{
\begin{tabular}{ccccc|c}
\toprule
LiDAR & ADA-Fusion & LDFA & Smoothing & Mamba head & mIoU         \\ 
\midrule
     &  & & & & 35.5   \\ 
     \checkmark& & & & & 44.4   \\ 
     \checkmark& \checkmark&  & & & 46.2 \\ 
     \checkmark& \checkmark& \checkmark& & & 48.3   \\
     \checkmark& \checkmark& \checkmark& \checkmark & & 48.7   \\
     \checkmark& \checkmark& \checkmark& \checkmark& \checkmark& \textbf{49.4}   \\
\bottomrule
\end{tabular}
}
\caption[]{Ablation study of different components on the Occ3D validation set. The baseline is the camera only GaussianFormer. We incrementally integrate our proposed modules to validate their effectiveness.}
\label{table:abb}
\end{table}

\subsection{Performance Analysis on Different Weather Conditions}

To examine the safety-critical robustness of the proposed framework, an evaluation is performed with the validation set of the nuScenes~\cite{nuscenes} dataset described under rainy conditions. Adverse weather, specifically rain, is a tough environmental setting for any perception solution. Raindrops casting refractions or occlusions over lenses of cameras, and wet surfaces that reflect like mirrors, are challenges for semantic understanding. It is worth noting that reduced ambient lighting, often accompanying rainy conditions, exacerbates the signal-to-noise ratio of visual sensors. Results of this study are described in Table~\ref{occ_rainy}.

\textbf{Limitations of Vision-Centric Approaches:} The susceptibility of camera-only architectures in this regard is evident. As shown in the first block of Table~\ref{occ_rainy}, the mIoU metric of state-of-the-art vision-based approaches like OccFusion (Camera-only) and SurroundOcc~\cite{surroundocc} is significantly low, at just 18.9\% and 19.8\%, respectively. This is indicative of a critical drawback since, in a scenario wherein the dominant cue or channel, photometric intensity, is disrupted by environmental noise, a model not possessing geometric redundancy is not capable of acquiring a correct representation of the scene geometry.

\textbf{Impact of Multi-Modal Fusion:} The introduction of LiDAR results in an instant improvement in accuracy. Comparing the mIoU value for the camera-only version of OccFusion~\cite{occfusionming2024} with the multi-modal version, OccFusion~\cite{occfusionming2024} C\&L, results reveal that mIoU improves from 18.9\% to 26.5\%. The implication of this result is that for accurate perception, even during adverse weather conditions, methods that are not hampered by changes in environmental brightness, such as LiDAR, have a crucial role.

\textbf{Superiority of the Proposed Framework:} Our GaussianOcc3D model has shown better performance in comparison to the hard multimodal baseline, OccFusion C\&L, by achieving the best mean intersection over union of 27.1\%. One can note that although both models make use of the same set of sensors, the better performance of our model can be attributed to the Adptive Camera-LiDAR Fusion component and the entropy-based smoothing component in the virtual camera component. The presence of rain in the image can introduce high entropy areas, including the blurry effect on pedestrians and road areas. However, unlike traditional concatenation-based techniques which may spread the effects in the occupancy map, the entropy-based smoothing component of the virtual camera is capable of detecting the inconsistencies in the image and smoothing out the less reliable areas before making the final occupation map using the most reliable sensor data, which in this case would be the LiDAR map, thanks to the adaptive fusion gate.

\definecolor{barriercolor}{RGB}{250, 141, 52}
\definecolor{bicyclecolor}{RGB}{245, 150, 100}
\definecolor{buscolor}{RGB}{250, 80, 100}
\definecolor{carcolor}{RGB}{255, 158, 0}
\definecolor{constcolor}{RGB}{230, 230, 0} 
\definecolor{motorcolor}{RGB}{255, 99, 71} 
\definecolor{pedcolor}{RGB}{0, 0, 230}     
\definecolor{conecolor}{RGB}{240, 60, 0}   
\definecolor{trailercolor}{RGB}{255, 140, 0}
\definecolor{truckcolor}{RGB}{255, 127, 80}
\definecolor{drivecolor}{RGB}{160, 60, 60} 
\definecolor{flatcolor}{RGB}{175, 255, 46} 
\definecolor{sidecolor}{RGB}{75, 0, 75}    
\definecolor{terraincolor}{RGB}{150, 240, 80}
\definecolor{mancolor}{RGB}{230, 230, 250} 
\definecolor{vegcolor}{RGB}{0, 175, 0}     

\begin{table}[htbp]
\centering
\fontsize{8pt}{10pt}\selectfont
\setlength{\tabcolsep}{0.4mm} 
\begin{threeparttable}
\begin{tabular}{l|c|c|cccccccccccccccc} 
\toprule

Method & Modality & mIoU & 
\makecell[b]{\begin{turn}{90}\colorbox{barriercolor}{\hspace{3pt}\rule{0pt}{3pt}} barrier\end{turn}} & 
\makecell[b]{\begin{turn}{90}\colorbox{bicyclecolor}{\hspace{3pt}\rule{0pt}{3pt}} bicycle\end{turn}} & 
\makecell[b]{\begin{turn}{90}\colorbox{buscolor}{\hspace{3pt}\rule{0pt}{3pt}} bus\end{turn}} & 
\makecell[b]{\begin{turn}{90}\colorbox{carcolor}{\hspace{3pt}\rule{0pt}{3pt}} car\end{turn}} & 
\makecell[b]{\begin{turn}{90}\colorbox{constcolor}{\hspace{3pt}\rule{0pt}{3pt}} const. veh.\end{turn}} & 
\makecell[b]{\begin{turn}{90}\colorbox{motorcolor}{\hspace{3pt}\rule{0pt}{3pt}} motorcycle\end{turn}} & 
\makecell[b]{\begin{turn}{90}\colorbox{pedcolor}{\hspace{3pt}\rule{0pt}{3pt}} pedestrian\end{turn}} & 
\makecell[b]{\begin{turn}{90}\colorbox{conecolor}{\hspace{3pt}\rule{0pt}{3pt}} traffic cone\end{turn}} & 
\makecell[b]{\begin{turn}{90}\colorbox{trailercolor}{\hspace{3pt}\rule{0pt}{3pt}} trailer\end{turn}} & 
\makecell[b]{\begin{turn}{90}\colorbox{truckcolor}{\hspace{3pt}\rule{0pt}{3pt}} truck\end{turn}} & 
\makecell[b]{\begin{turn}{90}\colorbox{drivecolor}{\hspace{3pt}\rule{0pt}{3pt}} drive. surf.\end{turn}} & 
\makecell[b]{\begin{turn}{90}\colorbox{flatcolor}{\hspace{3pt}\rule{0pt}{3pt}} other flat\end{turn}} & 
\makecell[b]{\begin{turn}{90}\colorbox{sidecolor}{\hspace{3pt}\rule{0pt}{3pt}} sidewalk\end{turn}} & 
\makecell[b]{\begin{turn}{90}\colorbox{terraincolor}{\hspace{3pt}\rule{0pt}{3pt}} terrain\end{turn}} & 
\makecell[b]{\begin{turn}{90}\colorbox{mancolor}{\hspace{3pt}\rule{0pt}{3pt}} manmade\end{turn}} & 
\makecell[b]{\begin{turn}{90}\colorbox{vegcolor}{\hspace{3pt}\rule{0pt}{3pt}} vegetation\end{turn}} \\

\midrule

OccFusion~\cite{occfusionming2024} & C  & 18.9 & 18.5 & \underline{14.2} & 22.2 & 30.0 & 10.1 & 15.2 & 10.0 & 9.7 & 13.2 & 20.9 & 37.1 & 23.4 & 27.7 & 17.4 & 10.3 & 23.1 \\

SurroundOcc~\cite{surroundocc} & C  & 19.8 & 21.4 & 12.7 & 25.4 & 31.3 & 11.3 & 12.6 & 8.9 & 9.4 & 14.5 & 21.5 & 35.3 & \textbf{25.3} & 29.8 & 18.3 & 14.4 & 24.7 \\

\midrule

OccFusion~\cite{occfusionming2024} &  C\&L  & \underline{26.5} & \underline{24.9} & 19.1 & \textbf{34.2} & \underline{36.0} & \underline{17.0} & \underline{21.0} & \underline{18.8} & \textbf{17.4} & \textbf{21.8} & \underline{28.7} & \underline{37.8} & \underline{24.3} & \underline{30.8} & \textbf{20.3} & \underline{28.9} & \underline{43.1} \\

\rowcolor{lightgray!45}  \textbf{GaussianOcc3D (ours)} &  C\&L  & \textbf{27.1} & \textbf{25.8} & \textbf{19.6} & \underline{34.1} & \textbf{36.4} & \textbf{18.1} & \textbf{21.3} & \textbf{19.9} & \underline{17.3} & \underline{21.6} & \textbf{29.5} & \textbf{38.2} & 24.1 & \textbf{30.9} & \underline{20.0} & \textbf{29.4} & \textbf{45.2} \\

\bottomrule
\end{tabular}
\end{threeparttable}
\caption[]{\textbf{3D semantic occupancy prediction results on nuScenes validation rainy scenario subset}.}
\label{occ_rainy}
\end{table}

\textbf{Night Condition:} Besides testing the performance in adverse weather conditions, the approach was also evaluated in extreme low-light conditions through the use of the night scenario validation subset in the nuScenes~\cite{nuscenes} dataset. Night driving presents a specific problem to the task of multi-modal perception. Unlike rainy situations where the visual information could be corrupted, in night situations, the absence of dynamic range or the degradation of the visual information in the areas not illuminated by the streetlights has often been found to be complete. The results of the evaluation are given in Table~\ref{occ_night}.

Besides testing the performance in adverse weather conditions, our approach was also evaluated in extreme low-light conditions through the use of the night scenario validation subset in the nuScenes~\cite{nuscenes} dataset. Night driving presents a unique challenge to multi-modal perception; unlike rainy conditions where visual information is merely corrupted, night scenes often suffer from a complete absence of dynamic range in areas not illuminated by streetlights or headlights. 

The results of this evaluation are detailed in Table~\ref{occ_night}. Qualitatively, this robust performance is demonstrated in Figure~\ref{fig:night}, which showcases samples from the OpenOccupancy validation set. As visualized, our model successfully reconstructs the 3D scene geometry despite the degraded visual input. The adaptive fusion mechanism effectively leverages LiDAR data in pitch-black regions while preserving fine-grained textural details in areas captured by the camera's dynamic range, such as those within the ego-vehicle's headlight beam. This balance allows for a consistent occupancy representation where baseline methods typically fail due to high-entropy visual noise.

\begin{table}[htbp]
\centering
\fontsize{8pt}{10pt}\selectfont
\setlength{\tabcolsep}{0.4mm} 
\begin{threeparttable}
\begin{tabular}{l|c|c|cccccccccccccccc} 
\toprule

Method & Modality & mIoU & 
\makecell[b]{\begin{turn}{90}\colorbox{barriercolor}{\hspace{3pt}\rule{0pt}{3pt}} barrier\end{turn}} & 
\makecell[b]{\begin{turn}{90}\colorbox{bicyclecolor}{\hspace{3pt}\rule{0pt}{3pt}} bicycle\end{turn}} & 
\makecell[b]{\begin{turn}{90}\colorbox{buscolor}{\hspace{3pt}\rule{0pt}{3pt}} bus\end{turn}} & 
\makecell[b]{\begin{turn}{90}\colorbox{carcolor}{\hspace{3pt}\rule{0pt}{3pt}} car\end{turn}} & 
\makecell[b]{\begin{turn}{90}\colorbox{constcolor}{\hspace{3pt}\rule{0pt}{3pt}} const. veh.\end{turn}} & 
\makecell[b]{\begin{turn}{90}\colorbox{motorcolor}{\hspace{3pt}\rule{0pt}{3pt}} motorcycle\end{turn}} & 
\makecell[b]{\begin{turn}{90}\colorbox{pedcolor}{\hspace{3pt}\rule{0pt}{3pt}} pedestrian\end{turn}} & 
\makecell[b]{\begin{turn}{90}\colorbox{conecolor}{\hspace{3pt}\rule{0pt}{3pt}} traffic cone\end{turn}} & 
\makecell[b]{\begin{turn}{90}\colorbox{trailercolor}{\hspace{3pt}\rule{0pt}{3pt}} trailer\end{turn}} & 
\makecell[b]{\begin{turn}{90}\colorbox{truckcolor}{\hspace{3pt}\rule{0pt}{3pt}} truck\end{turn}} & 
\makecell[b]{\begin{turn}{90}\colorbox{drivecolor}{\hspace{3pt}\rule{0pt}{3pt}} drive. surf.\end{turn}} & 
\makecell[b]{\begin{turn}{90}\colorbox{flatcolor}{\hspace{3pt}\rule{0pt}{3pt}} other flat\end{turn}} & 
\makecell[b]{\begin{turn}{90}\colorbox{sidecolor}{\hspace{3pt}\rule{0pt}{3pt}} sidewalk\end{turn}} & 
\makecell[b]{\begin{turn}{90}\colorbox{terraincolor}{\hspace{3pt}\rule{0pt}{3pt}} terrain\end{turn}} & 
\makecell[b]{\begin{turn}{90}\colorbox{mancolor}{\hspace{3pt}\rule{0pt}{3pt}} manmade\end{turn}} & 
\makecell[b]{\begin{turn}{90}\colorbox{vegcolor}{\hspace{3pt}\rule{0pt}{3pt}} vegetation\end{turn}} \\

\midrule

OccFusion~\cite{occfusionming2024} & C  & 9.99 & 10.40 & 12.03 & 0.00 & 29.94 & 0.00 & 9.92 & 4.88 & \underline{0.91} & 0.00 & 17.79 & 29.10 & \underline{2.37} & 10.80 & 9.40 & 8.68 & 13.57 \\

SurroundOcc~\cite{surroundocc} & C  & 10.80 & 10.55 & \underline{14.60} & 0.00 & 31.05 & 0.00 & 8.26 & 5.37 & 0.58 & 0.00 & 18.75 & 30.72 & \textbf{2.74} & 12.39 & 11.53 & 10.52 & 15.77 \\

\midrule

OccFusion~\cite{occfusionming2024} & C\&L  & \underline{15.2} & \underline{12.7} & 13.5 & 0.0 & \underline{35.8} & 0.0 & \textbf{15.3} & \underline{13.1} & 0.8 & 0.0 & \textbf{23.7} & \underline{32.4} & 0.9 & \underline{14.2} & \textbf{20.5} & \textbf{23.5} & \underline{37.1} \\

\rowcolor{lightgray!45}  \textbf{GaussianOcc3D (ours)} & C\&L  & \textbf{15.9} & \textbf{13.9} & \textbf{14.8} & 0.0 & \textbf{35.9} & 0.0 & \underline{15.1} & \textbf{14.5} & \textbf{1.5} & 0.0 & \underline{23.4} & \textbf{34.7} & 1.6 & \textbf{15.3} & \underline{20.4} & \underline{23.2} & \textbf{39.3} \\

\bottomrule
\end{tabular}
\end{threeparttable}
\caption[]{3D semantic occupancy prediction results on nuScenes validation night scenario subset.}
\label{occ_night}
\end{table}

\textbf{Failure of Vision-Centric Baselines:} The outcome illustrates the importance of photometric information in standard occupancy networks. As shown in the early part of the table, camera-only approaches show a dramatically decreased performance. To be more precise, OccFusion~\cite{occfusionming2024} (Camera-only) and SurroundOcc~\cite{surroundocc} achieve mean intersection over union (mIoU) metrics of 9.99\% and 10.80\%, respectively. These results come near to random guesswork on most semantic segments, thus validating the importance of passive sensors in ensuring the safety of 3D environment reconstruction in the absence of ambient illumination.

\textbf{The Critical Role of Active Sensing:} The combination of Light Detection and Ranging (LiDAR) technology, a form of active sensor that provides illumination, is in fact the main reason behind the recovery of performance in these settings. The multi-modal baseline, OccFusion~\cite{occfusionming2024} C\&L, exhibits a large relative gain, reaching a result of 15.2\% mIoU. This verifies the assertion that geometrical priors are not simply supplementals but essential in low-light conditions, providing the structure that is not possible by cameras only.

\textbf{Efficacy of Adaptive Fusion in Extreme Darkness:} Our GaussianOcc3D model performs best, scoring an mIoU of 15.9\%, outperforming the multimodal baseline model. Even though a difference of a few percent might not appear dramatic, this is a large improvement considering that this is a very challenging subset, with several classes registering 0.00\% in both approaches because of the sparsity of their representation. This is, of course, a direct result of our Adaptive Camera-LiDAR Fusion module. When in completely dark areas, there is a lot of high entropy noise from the camera modality. The specific gating network in our fusion module is responsible for suppressing this type of visual information, which instead puts more weight on the LiDAR modality. In areas illuminated by headlights, like the one on a driving path in front of the ego-vehicle, the gate lets through visual information about texture.

\subsection{Performance Analysis on Number of Gaussians}

\begin{figure}[ht]
    \centering
    \includegraphics[width=1\textwidth]{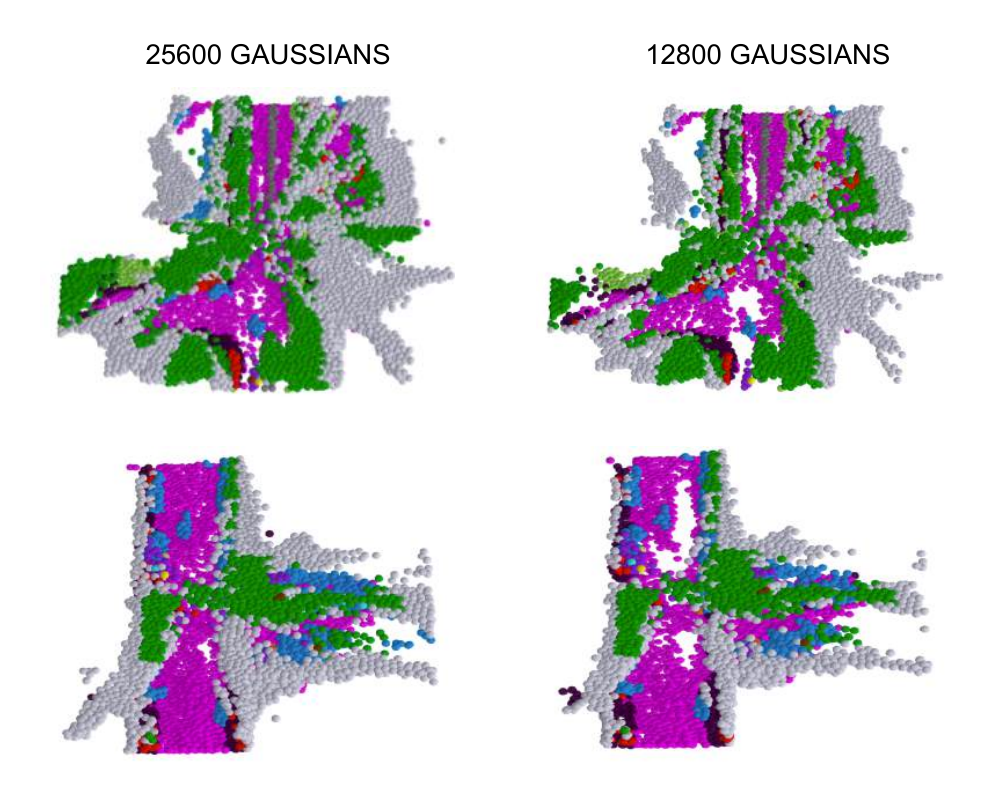} 
    \caption[]{Comparison of 12,800 and 25,600 Gaussian representations on the OpenOccupancy validation set. }
    \label{fig:gaussians}
\end{figure}

To demonstrate the effect of representation density on the quality of scene reconstruction, an ablation study was performed by controlling the number of Gaussian primitives used for representing a scene. Table~\ref{table:gaussian} presents the quantitative results on the OpenOccupancy validation set (v0.0 annotations) using 12,800 and 25,600 Gaussian primitives.

Analysis of the results shows a strong positive correlation between the number of primitives and the mean Intersection-over-Union (mIoU) value. Increasing the number of Gaussians from 12,800 to 25,600 leads to a performance gain of $+2.5$ percentage points ($22.8\% \rightarrow 25.3\%$). This improvement signifies that with higher density, the model is better equipped to identify smaller details and complex geometries within the scene.

As visualized in Figure~\ref{fig:gaussians}, the configuration with 25,600 Gaussians provides a much higher-fidelity reconstruction. While the 12,800 configuration exhibits visible effects of sparsity, the 25,600 density provides sufficient coverage to capture both large static background elements and slender foreground objects accurately. Based on these results, we set 25,600 Gaussians as the baseline for our final model due to the optimal trade-off between representation accuracy and computational efficiency.

\begin{table}[htbp]

\centering
\scalebox{1}{
\begin{tabular}{c|c}
\toprule
Number of Gaussians  & mIoU         \\ 
\midrule
      12800& 22.8  \\ 
      25600& \textbf{25.3} \\ 
\bottomrule
\end{tabular}
}
\caption[]{Performance analysis of different numbers of gaussians on the OpenOccupancy validation set.}
\label{table:gaussian}
\end{table}

\subsection{Performance Analysis of Different Fusion Techniques}

In order to substantiate the validity of the designed architecture for the Adaptive Camera-LiDAR Fusion (ADA-Fusion) module, a comparison has been done between the proposed architecture and the general feature aggregation operations used in the literature. This validation has been done through the use of Table~\ref{table:fusion}.

\textbf{Limitations of Static Fusion:} We began by evaluating Element-Wise Addition. This had the weakest performance and an mIOU score of 47.3\%. This operation may be computationally efficient but tends to have a fixed combination of the feature spaces. This often leads to destructive interference. For example, a large value of noise in the modality (such as glare in the image projected by the camera) may interfere with the true signal in the other modality and hence cause the signal-to-noise ratio to drop.

\textbf{Concatenation vs. Addition:} The standard concatenation operation, forming the main baseline for most multimodal approaches, achieves an mIoU of 47.8\% (improving over the addition baseline by 0.5\%). It does this by maintaining the same dimensionality of the channels in both modalities. Therefore, the convolutional layers are able to learn different weights for the visual and geometric representations. However, this is still an operation that does not change. The weights learned are applied uniformly through the input image and thus do not represent the varying reliability of the sensors (e.g., where the camera may be occluded and the LiDAR sensor may not be).

\textbf{Superiority of ADA-Fusion:} The ADA-Fusion mechanism demonstrates a significant performance gain with a result of 49.4\% mIoU. This presents a significant improvement of 1.6\% over the robust concatenation method. The improvement is directly related to the dynamic characteristics of the ADA-Fusion mechanism. Unlike the static method of concatenation, ADA-Fusion utilizes the Soft Gating and Consistency-Aware Reweighting method to dynamically evaluate the consistency of the sensors per-point. By down-weighting the channel of the features when the sensors disagree and up-weighting when the sensors agree, the ADA-Fusion mechanism overcomes the negative transfer learning effect, which leads to the creation of a cleaner representation than the static method of concatenation.

\begin{table}[htbp]
\centering
\scalebox{1}{
\begin{tabular}{ccc|c}
\toprule
Addition & Concatenation & ADA-Fusion& mIoU         \\ 
\midrule
     \checkmark&  & & 47.3  \\ 
     & \checkmark&  & 47.8  \\ 
     & &  \checkmark& \textbf{49.4}  \\ 
\bottomrule
\end{tabular}
}
\caption[]{Performance analysis of different fusion techniques on the Occ3D validation set.}
\label{table:fusion}
\end{table}

\subsection{Analysis of Complexity}

To assess the computational efficiency of the proposed framework, we compare the number of learnable parameters and inference latency against representative state-of-the-art methods. Table~\ref{table:complexity} summarizes these metrics.

\begin{table}[htbp]
\centering
\scalebox{0.8}{
\begin{tabular}{c|c|c|c}
\toprule
Method  &  Modality & Number of Params. (M) & Latency (ms)         \\ 
\midrule
       GaussianFormer~\cite{huang2024gaussianformer} & C &52.4  &1280  \\ 
       \midrule
      OccMamba~\cite{occmamba2024}& C\&L &92.3 & 1820  \\
      \textbf{GaussianOcc3D (ours)} & C\&L &68.1 &1650 \\ 
\bottomrule
\end{tabular}
}
\caption[]{Analysis of different model complexity}
\label{table:complexity}
\end{table}

Compared to the state-of-the-art multi-modal baseline OccMamba~\cite{occmamba2024}, the GaussianOcc3D framework shows a significant improvement in terms of parameter usage. For instance, the OccMamba framework uses 92.3 million parameters to analyze the blended camera and LiDAR inputs. On the other hand, the proposed GaussianOcc3D framework achieves better results using only 68.1 million parameters, a drop of nearly 26\% from the previous approach. This is because the new GaussianOcc3D framework eliminates the complex voxel reconstruction and uses the linear complexity Gauss-Mamba head.

In terms of runtime, our method achieves an inference latency of 1650 ms, offering a distinct speed advantage over OccMamba (1820 ms). Although the camera-only GaussianFormer~\cite{huang2024gaussianformer} remains the lightest architecture (52.4 M parameters, 1280 ms), this is an expected trade-off given its lack of a LiDAR backbone. Crucially, GaussianOcc3D successfully integrates a second sensory modality with only a moderate increase in computational cost, striking a favorable balance between high-fidelity multi-modal perception and practical deployment feasibility.

\subsection{Qualitative Results}

\begin{figure}[ht]
    \centering
    \includegraphics[width=1\textwidth]{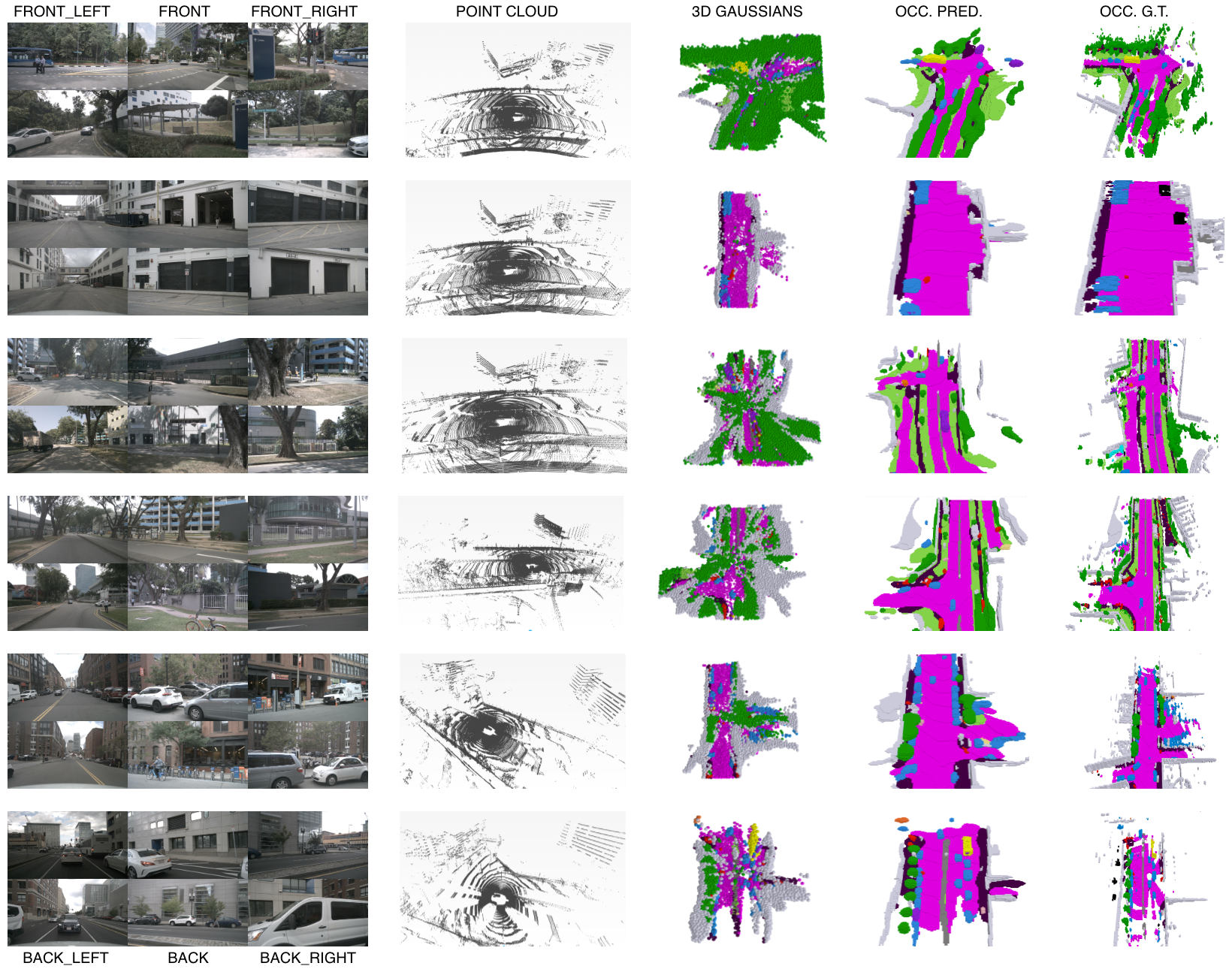} 
    \caption[]{Qualitative results on the OpenOccupancy validation set.}
    \label{fig:daily}
\end{figure}

To evaluate the robustness of our framework across diverse real-world environments, we present qualitative results across three distinct lighting and weather conditions. 
As illustrated in Figure~\ref{fig:daily}, the daily samples demonstrate the model's high-fidelity reconstruction capabilities under optimal illumination. In these scenarios, the predicted occupancy (OCC. PRED.) closely mirrors the ground truth (OCC. G.T.), accurately capturing complex urban structures and intricate road layouts. 

\begin{figure}[ht]
    \centering
    \includegraphics[width=1\textwidth]{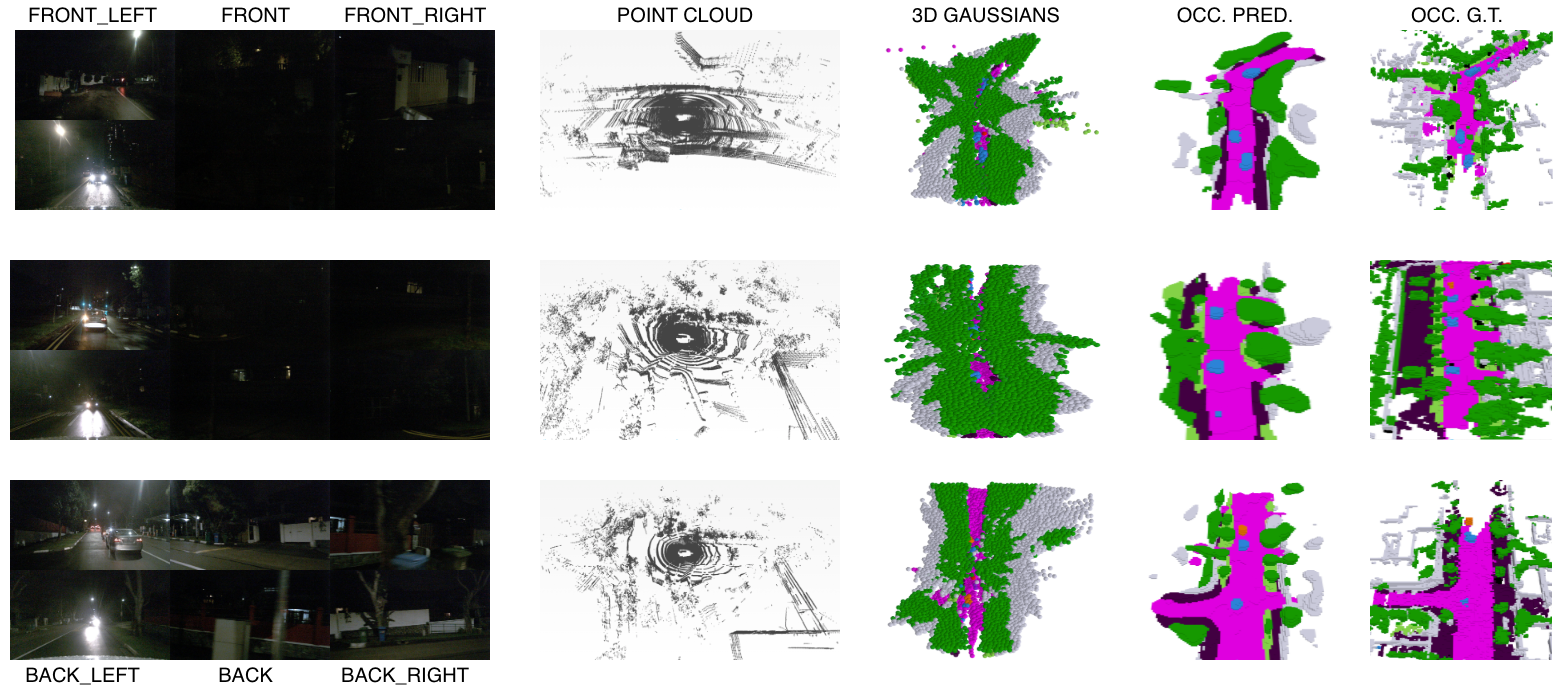} 
    \caption[]{Qualitative results on the OpenOccupancy validation set under night-time conditions. }
    \label{fig:night}
\end{figure}

The performance under night conditions is presented in Figure~\ref{fig:night}, where camera visibility is significantly constrained by low-light environments and high-contrast glare. Despite these sensory limitations, our approach leverages the geometric consistency of the 3D Gaussians to maintain a reliable occupancy map, proving its efficacy for safety-critical nighttime navigation. 

\begin{figure}[ht]
    \centering
    \includegraphics[width=1\textwidth]{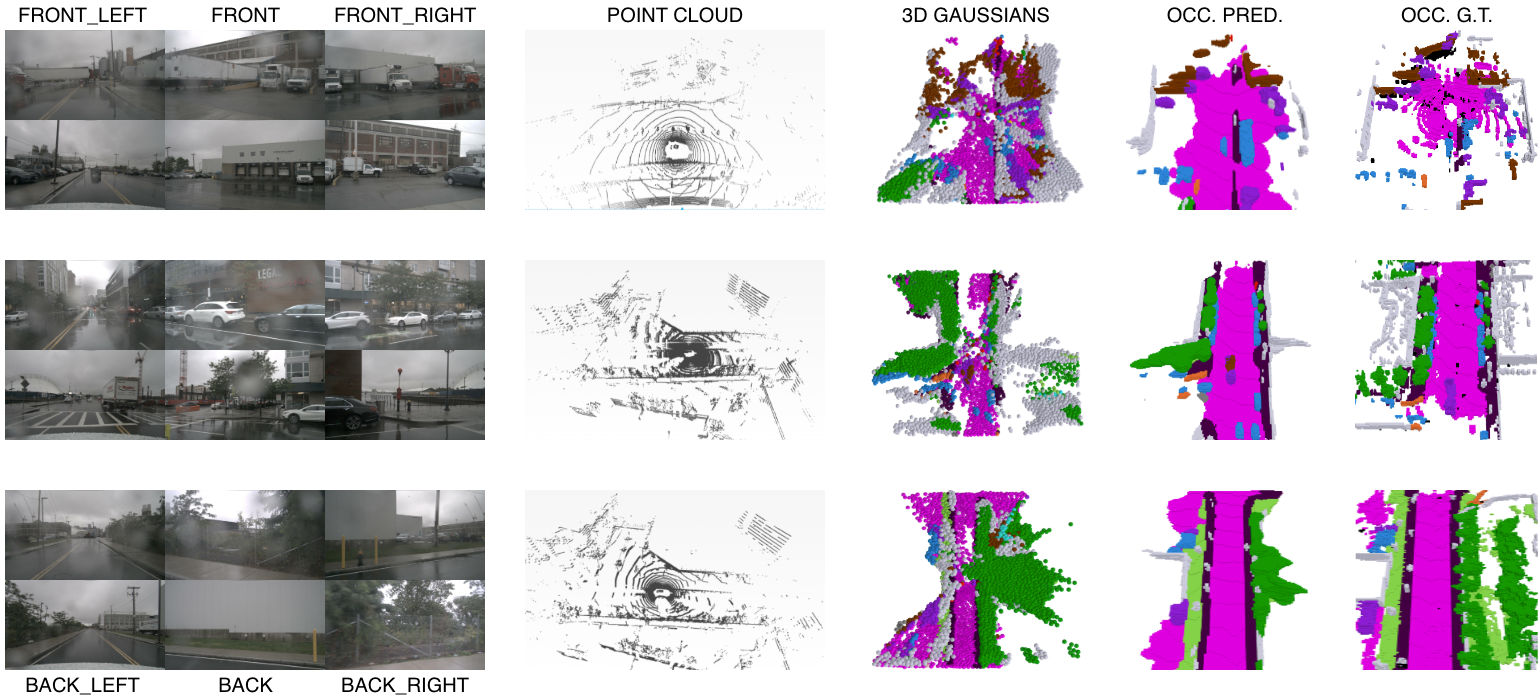} 
    \caption[]{Qualitative results on the OpenOccupancy validation set under rainy conditions. }
    \label{fig:rainy}
\end{figure}

Finally, Figure~\ref{fig:rainy} displays results from rainy conditions, which are characterized by visual noise, precipitation artifacts, and reflections on the camera lenses. The framework demonstrates impressive resilience to these environmental disturbances, producing stable and coherent spatial representations that align closely with the underlying point cloud data. Collectively, these visualizations confirm that the proposed method generalizes effectively across varying atmospheric and temporal conditions, providing a consistent 360-degree understanding of the surrounding scene.

\chapter{Conclusions}


The prime objective of the thesis at hand is to investigate the most significant problem related to 3D semantic occupancy prediction in the context of autonomous driving. These problems are the interplay between the resolution of representation and the efficiency of computation as well as the handling of the heterogeneities of various sensor modalities. Owing to the advancement of autonomous systems from sparse object detection to dense scene understanding, the shortcoming of traditional voxel representations in the form of grids—due to their cubic memory requirement and the effect of quantization artifacts—has become an important limitation. At the same time, traditional fusion approaches often rely on fixed models that are unable to convey the varying accuracy of camera and LiDAR sensors.


To counter the said limitations, the thesis proposes the Gaussian-Based Adaptive Camera-LIDAR Multimodal 3D Occupancy Prediction solution. By moving the paradigm of representation from discrete voxels to continuous and anisotropic three-dimensional Gaussians, we have successfully decoupled the scene resolution and the memory requirements. The proposed methodology incorporates the following newly added components:

\begin{itemize}
    \item \textbf{LiDAR Depth Feature Aggregation (LDFA):} A module that leverages depth-wise deformable sampling to resolve the sparsity of geometric data.
    \item \textbf{Entropy-Based Feature Smoothing:} A stochastic regularization technique capable of mitigating domain-specific noise and preventing negative transfer between conflicting sensor streams.
    \item \textbf{Adaptive Camera–LiDAR Fusion:} A dynamic fusion engine that utilizes uncertainty-aware reweighting to prioritize the most reliable sensor modality in real-time.
    \item \textbf{The Gauss-Mamba Head:} An efficient decoding architecture that applies Selective State Space Models (Mamba) within a Tri-Perspective View decomposition to capture global context with linear complexity.
\end{itemize}

\section{Key Findings and Contributions}

The experimental validation conducted across three diverse and large-scale benchmarks—OpenOccupancy (nuScenes), Occ3D-nuScenes, and SemanticKITTI—confirms the efficacy of the proposed framework. The quantitative results demonstrate that our approach consistently outperforms state-of-the-art baselines, achieving mean Intersection-over-Union (mIoU) scores of 25.3\%, 49.4\%, and 25.2\%, respectively.

Beyond aggregate metrics, a granular analysis of the results yields several critical insights:
\begin{enumerate}
    \item \textbf{Geometric Fidelity of Thin Structures:} The framework exhibited superior performance on classes defined by thin, irregular geometries, such as Bicycles, Barriers, and Vegetation. This validates the hypothesis that anisotropic Gaussian primitives are fundamentally better suited for modeling non-cubic structures than rigid voxel grids, effectively reducing quantization errors.
    \sloppy
    \item \textbf{Robustness in Adverse Conditions:} The adaptive fusion mechanism demonstrated significant resilience in challenging subsets, such as rain and night scenarios. By dynamically down-weighting unreliable photometric signals in low-light conditions or noisy LiDAR returns during precipitation, the model maintained high perceptual integrity where static fusion methods degraded.
    \item \textbf{Efficiency via Mamba:} The integration of the Gauss-Mamba head proved that global contextual reasoning can be achieved without the quadratic computational cost of Transformers, enabling high-resolution predictions suitable for real-time applications.
\end{enumerate}

\section{Limitations}

Despite the promising performance, several limitations of the current framework warrant acknowledgement:
\begin{itemize}
    \item \textbf{Calibration Sensitivity:} As with all explicit lifting strategies, the proposed method is sensitive to the accuracy of extrinsic calibration between the camera and LiDAR. Slight misalignments can lead to feature smearing during the Gaussian lifting process.
    \item \textbf{Temporal Consistency:} The current implementation processes scenes primarily in a frame-by-frame manner. While the Mamba head captures spatial context, the lack of explicit temporal propagation means that flickering artifacts may occasionally occur in the occupancy predictions of moving objects over time.
\end{itemize}

\section{Future Work}

Building upon the foundations laid by this thesis, several avenues for future research emerge:

\begin{itemize}
    \item \textbf{4D Spatiotemporal Gaussian Splatting:} Extending the framework to the temporal domain (4D) could significantly improve consistency. Future work could explore tracking Gaussian primitives across frames, allowing the model to leverage historical context to resolve occlusions and smooth predictions for dynamic agents.
    \item \textbf{End-to-End Planning Integration:} Currently, the occupancy grid serves as an intermediate representation. Integrating this Gaussian-based perception directly with downstream planning and control modules could enable an end-to-end learning pipeline, where the occupancy prediction is optimized specifically for navigability and safety rather than just semantic classification.
    \item \textbf{Self-Supervised Adaptation:} To further reduce the reliance on expensive dense occupancy annotations, future iterations could incorporate self-supervised learning objectives, utilizing photometrical rendering losses (rendering the Gaussians back to 2D images) to fine-tune the model on unlabeled data.
    \item \textbf{Edge Deployment Optimization:} While efficient, the current model relies on high-end GPUs. optimizing the Gauss-Mamba kernels for embedded hardware (e.g., NVIDIA Jetson or Drive Orin platforms) is a necessary step for commercial deployment in production vehicles.
\end{itemize}

In conclusion, this thesis represents a significant step forward in multi-modal 3D perception. By harmonizing the representational flexibility of 3D Gaussians with the computational efficiency of State Space Models, we have established a robust, scalable, and accurate framework for the next generation of autonomous driving systems.

\begin{postliminary}
\references

\newpage
\vspace*{\stretch{1}}

\chapter{vita}

Abdullah Enes Doruk received his Bachelor’s degree in Electrical and Electronics Engineering from Bursa Technical University in 2022. During his undergraduate studies, he gained experience through internships and participated in several TEKNOFEST competitions in the fields of AI in transportation, unmanned systems, and autonomous vehicles. In 2022, he began working as an AI Robotic Software Engineer at DAIMIA Engineering, followed by a role as a Perception and Detection Software Engineer at ADASTEC Corporation in 2023. He also contributed to a TÜBİTAK 2209-B industry-oriented research project focusing on applied machine learning. In 2023, he started his Master’s degree in Artificial Intelligence Engineering at Özyeğin University, where he has been working as a Graduate Researcher at the DeepVIP Lab under the supervision of Prof. Dr. Hasan Fehmi Ates. His research interests include 3D computer vision, unsupervised domain adaptation, and multi-sensor fusion for autonomous driving and intelligent perception systems.

\end{postliminary}

\end{document}